\documentclass[lettersize,journal]{IEEEtran}
\usepackage{textcomp}
\usepackage{verbatim}
\usepackage{graphicx}
\usepackage[T1]{fontenc}
\usepackage{mathptmx}
\usepackage{amsmath}
\usepackage{amsfonts}
\usepackage{amssymb}
\usepackage{graphicx}
\usepackage{algorithm}
\usepackage{algorithmicx}
\usepackage{algpseudocode}
\usepackage{booktabs}
\usepackage{array}
\usepackage{multirow}
\usepackage{xspace}
\usepackage{xcolor}
\usepackage{colortbl}
\usepackage{makecell}
\usepackage{subfigure}
\usepackage{CJK}
\usepackage{float}
\usepackage{balance}
\usepackage[pagebackref=false,breaklinks=false,letterpaper=false,colorlinks,bookmarks=false]{hyperref}
\usepackage{url}
\newcommand{\ie}{i.e.,\ }

\newcommand{\et}{\emph{et al.}\ }

\begin{document}

\title{In Defense and Revival of Bayesian Filtering for Thermal Infrared Object Tracking}

\author{
Peng Gao,
Shi-Min Li,
Feng Gao,
Fei~Wang,
Ru-Yue Yuan,
Hamido Fujita

\thanks{P. Gao and S.-M. Li are with the School of Cyber Science and Engineering, Qufu Normal University, Qufu, Shandong 273165, China.}
\thanks{F. Gao is with School of Computer Science and Technology, East China Normal University, Shanghai 200062, China.}
\thanks{F. Wang is School of Integrated Circuits, Harbin Institute of Technology, Shenzhen, Guangdong 518055, China.}
\thanks{R.-Y. Yuan is an individual researcher.}
\thanks{H. Fujita is with the Malaysia-Japan International Institute of Technology (MJIIT), Universiti Teknologi Malaysia, Kuala Lumpur, 54100, Malaysia, and also with the Faculty of Software and Information Science, Iwate Prefectural University, Takizawa, Iwate 020-0611, Japan.}
}

\maketitle

\begin{abstract}
Deep learning-based methods monopolize the latest research in the field of thermal infrared (TIR) object tracking. However, relying solely on deep learning models to obtain better tracking results requires carefully selecting feature information that is beneficial to representing the target object and designing a reasonable template update strategy, which undoubtedly increases the difficulty of model design. Thus, recent TIR tracking methods face many challenges in complex scenarios. This paper introduces a novel Deep Bayesian Filtering (DBF) method to enhance TIR tracking in these challenging situations. DBF is distinctive in its dual-model structure: the system and observation models. The system model leverages motion data to estimate the potential positions of the target object based on two-dimensional Brownian motion, thus generating a prior probability. Following this, the observation model comes into play upon capturing the TIR image. It serves as a classifier and employs infrared information to ascertain the likelihood of these estimated positions, creating a likelihood probability. According to the guidance of the two models, the position of the target object can be determined, and the template can be dynamically updated. Experimental analysis across several benchmark datasets reveals that DBF achieves competitive performance, surpassing most existing TIR tracking methods in complex scenarios.
\end{abstract}


\section{Introduction}\label{sec:introduction}

Thermal infrared (TIR) object tracking, a key issue in computer vision, has extensive applications in fields like human-computer interaction, video surveillance, unmanned vehicles, and motion analysis. This process entails using TIR cameras to estimating the position and size of a specified target object~\cite{felsberg2015thermal,yuan2022recent}. Unlike visible light cameras that capture light reflection, TIR cameras detect infrared radiation. Since all objects above absolute zero (-273.15 $^\circ$C or -459.67 $^\circ$F) emit infrared radiation, TIR imaging proves effective in complete darkness or challenging weather~\cite{sobrino2016review,gade2014thermal}. The advantages of TIR tracking include the ability to detect target objects in low-visibility conditions such as fog, smoke, or darkness, where visible light cameras might be ineffective. It also excels in identifying target objects with similar temperatures to their surroundings, a task difficult for other visual tracking methods~\cite{zhang2020object}.

Over the past decade, deep learning, particularly convolutional neural networks (CNNs), has emerged as a foundational element for feature extraction in image and video analysis, thereby improving a variety of computer vision tasks~\cite{matsuo2022deep,menghani2023efficient}. Among the diverse range of network architectures, Siamese networks have gained traction for their effectiveness in visual tracking~\cite{siamrpn,zhang2023target,satin,siamtpn,siamextr}. These networks function by matching a template of the target object against various candidate regions in subsequent frames, achieving efficient and robust performance~\cite{survey2022a}. Despite the advancements in deep learning for TIR tracking, challenges persist in certain complex scenarios~\cite{lsotb,ptb}, such as deformation (DEF), scale variation (SV), out-of-view (OV), thermal crossover (TC), motion blur (MB), and background clutter (BC). The constantly changing appearance and shape of the target object necessitate ongoing updates to the template~\cite{survey2021}. There are two principal strategies for this: (1) constant use the target object in the first frame as the template~\cite{siamfc,siamtri}, and (2) continuous updating of the template with each new frame~\cite{csot,eco,rar}. Nevertheless, these strategies can be problematic in intricate scenarios. The first strategy struggles with DEF, SV, and OV, as the unaltered template cannot adapt to the target object's evolving appearance and size, leading to tracking failures. The second strategy, while updating continuously, can be adversely affected by background distractions in cases of TC, MB, and BC, which results in error accumulation and tracking drift.

Recent developments in TIR tracking have largely focused on mitigating the limitations of existing strategies by either updating the template sporadically over several frames or adjusting the learning rate to partially update the template for each frame~\cite{mlssnet,yang2024learning}. This strategy, however, represents a compromise, often resulting in suboptimal performance in complex scenarios. The underlying issue, as we argue, is the exclusive reliance on a single type of data: infrared information is not sufficient on its own for effective TIR tracking in such contexts. The major developmental thrust in existing methods has been towards continuously enhancing the quality of critical representations, with the expectation that the template would become more adaptable to the diverse changes in the target object. Nevertheless, this line of approach is not adequate to fundamentally tackle the complexities of TIR tracking in challenging scenarios~\cite{mmnet}. To improve outcomes in these difficult situations, we contend, it's vital to integrate motion data that is independent of infrared information. By incorporating additional motion data, the tracker is better equipped to estimate the possible position of the target object, aiding the infrared information in effectively differentiating it from the background. For instance, in scenarios with BC, using motion data to forecast the target position can lead to more conservative updates to the template, thereby minimizing tracking errors. While some pioneering methods have incorporated optical flow as motion data~\cite{lmsco}, merging different types of features often requires elaborate network architectures and extensive training on larger datasets. This necessitates a more adaptive and holistic approach that skillfully combines the strengths of deep learning with additional data types.

Bayesian filtering, distinguished for its ability to proficiently handle uncertainty and integrate different data types, offers an effective solution to the previously outlined challenges~\cite{stano2013parametric,arulampalam2002tutorial}. Its effectiveness extends far beyond theoretical constructs, as evidenced by its successful implementation in multiple domains. In autonomous vehicle navigation, the use of Bayesian filters has proven essential for accurately determining vehicle positions and movements under uncertain conditions~\cite{fang2022inertial,huang2021variational}. Similarly, in robotics, Bayesian filters have significantly improved the ability of robots to navigate and interact in changing environments, thus enhancing their decision-making skills~\cite{choe2021indoor,dagan2022conservative}. In the field of signal processing, Bayesian methods have made considerable contributions, notably in noise reduction and signal improvement~\cite{petetin2021structured,martino2021compressed}. These diverse applications not only attest to the robustness and flexibility of Bayesian filtering but also underscore its potential to significantly refine TIR object tracking. By integrating Bayesian filtering, it becomes possible to provide motion information and combine it with infrared data for a more sophisticated and probabilistic analysis of the target position and movements, effectively overcoming the critical limitations of deep learning approaches in TIR tracking, especially in situations where infrared data is ambiguous or insufficient.

We believe incorporating Bayesian filtering into deep learning-based methods for TIR tracking yields two primary benefits. First, it enables the inclusion of motion data independent of infrared information, significantly enhancing the tracker's capacity to anticipate and discriminate to fast changes in the target position or appearance. This aspect is vital for ensuring tracking precision in scenarios where the infrared information alone may not suffice for accurate differentiation~\cite{bhat2019multi}. Second, Bayesian filtering introduces a dynamic model updating strategy, which often lacking in conventional deep learning techniques. Rather than relying on static or indiscriminate updating strategies, Bayesian filtering applies a nuanced approach, adjusting the model based on calculated probabilistic confidence~\cite{cao2021bayesian}. This method of adaptive updating greatly mitigates the risk of tracking drift and error accumulation, especially in scenarios characterized by intricate thermal behavior. Therefore, the integration of Bayesian filtering not only complements but also significantly enhances the performance of deep learning-based TIR tracking methods, fostering the development of more efficient and effective tracking technologies for a range of challenging situations.

Informed by our earlier analysis, we present a novel framework called Deep Bayesian Filtering (DBF), aimed at enhancing TIR tracking in complex scenarios through the efficient combination of motion and infrared information. DBF comprises two autonomous models: the system model and the observation model. The system model operates prior to the acquisition of the TIR image, leveraging motion data of the target object to estimate its possible positions (generating a prior probability), and is developed based on two-dimensional Brownian motion~\cite{iyengar1985hitting}. After the TIR image is captured, the observation model uses infrared information to determine the likelihood (generating likelihood probability) of the possible position, built on a deep learning-based tracking paradigm~\cite{mdnet}. These models converge using the Bayesian formula, which combines the prior probability distribution from the system model with the likelihood probability distribution from the observation model, culminating in the posterior probability distribution of the target position. The peak of this distribution serves as the estimated position.
Experimental results on several benchmark datasets have shown that DBF obtains competitive performance against most existing TIR tracking methods in complex scenarios. The significant contributions of this study can be outlined as follows:

\begin{itemize}
  \item We introduce Deep Bayesian Filtering, an innovative framework designed to merge motion and infrared information for enhanced TIR object tracking.
  \item The development of both a system model and an observation model, which together collaboratively predict the position of target object, is presented.
  \item The effectiveness of DBF is validated through comparative experiments on two prominent datasets, demonstrating its superiority over conventional trackers based on single infrared information.
\end{itemize}

The remainder of this paper is organized as follows. Recent advances related to TIR object tracking are introduced in Section~\ref{sec:2}. Section~\ref{sec:3} elaborates the methodology of the proposed DBF. Section~\ref{sec:4} details the experiments and discusses the results. Finally, a brief summary of this study is provided in Section~\ref{sec:5}.

\section{Related Work}\label{sec:2}

In this section, we outline relevant research. First, we explore existing methods in deep learning-based visual tracking in Section~\ref{sec:2-1}. Next, we examine recent representative TIR object tracking methods in Section~\ref{sec:2-2}. The section wraps up with an examination of how Bayesian approaches are used in other visual tasks in Section~\ref{sec:2-3}.

\subsection{Deep visual object tracking}\label{sec:2-1}

Deep learning has significantly transformed the landscape of visual object tracking, introducing unparalleled accuracy and robustness~\cite{survey2022}. This paradigm shift has catalyzed the development of sophisticated models capable of handling diverse and complex tracking scenarios. The integration of deep learning in object tracking began with works like DLT~\cite{dlt} and GOTURN~\cite{held2016learning}, which demonstrated the efficacy of CNNs. This marked a departure from traditional correlation filter-based tracking methods~\cite{csot,cact,mlcft,mfcmt}, paving the way for more advanced neural network applications. Subsequent advancements saw the emergence of various methods such as SiamFC~\cite{siamfc}, MDNet~\cite{mdnet}, SiamRPN~\cite{siamrpn}, and SiamEXTR~\cite{siamextr}. Each method offered unique strengths; for instance, Siamese networks gained popularity for their balance between accuracy and speed in real-time tracking applications~\cite{survey2021}. The development and refinement of these methods have been greatly influenced by large-scale training datasets~\cite{imagenet,coco,trackingnet}, and challenging benchmarks such as OTB~\cite{otb}, GOT-10k~\cite{got10k}, and LaSOT~\cite{lasot}. These resources have been crucial in evaluating the performance and robustness of tracking methods under varied conditions. Despite advancements, challenges such as OCC, SV, and BC remain persistent issues in deep learning-based trackers. Moreover, the generalizability of these methods across different environments and objects is an area of ongoing research. Recent notable studies introduce the famous Transformer~\cite{transformer} architecture to the visual tracking community, achieving state-of-the-art performance in most benchmark datasets~\cite{dropmae,ostrack,seqtrack}. Besides, in the broader context of visual object tracking, significant advancements have been made by segmentation-based methods~\cite{revision1,revision2,revision3}. While these methods focus on visible data, their underlying principles offer valuable insights into feature extraction, semantic segmentation, and integrating diverse data types. These studies reflect the continuous evolution and refinement of visual tracking methods.

\subsection{TIR object tracking}\label{sec:2-2}

TIR object tracking, leveraging TIR imaging to detect and follow objects, plays a critical role in scenarios with limited or no visible light. This tracking modality offers unique advantages over traditional visual tracking, particularly in challenging visibility conditions~\cite{yuan2022recent}. The integration of deep learning into TIR tracking has marked a significant advancement~\cite{liu2017deep}. Notable models such as CNNs and recurrent neural networks (RNNs)~\cite{schuster1997bidirectional} have been adapted to enhance the accuracy and reliability of TIR tracking methods. Integrating infrared information with other modalities, like RGB data, has been explored to improve robustness. This multimodal approach, as demonstrated in MMMPT~\cite{cai2023learning}, shows promise in creating more resilient tracking systems under varying conditions. Significant recent contributions include MLSSNet~\cite{mlssnet}, achieving breakthrough performance in distinguishing distractors, and MMNet~\cite{mmnet}, which innovated in learning dual-level deep representation. These studies showcase the rapid progress in TIR tracking. Emerging trends include the use of powerful Transformer architecture to further refine TIR tracking and the exploration of new feature-fusion technologies~\cite{huang2023rgb,qiu2023visible,li2023efficient}. It is also noteworthy that segmentation-based tracking methods have not yet been developed for TIR object tracking due to the lack of accurate pixel-level annotations. We believe TIR tracking has undergone substantial growth, primarily driven by advancements in deep learning. The ongoing research in this field is poised to significantly enhance capabilities in complex scenarios, expanding its applicability across various domains.

\subsection{Bayesian filtering in visual tasks}\label{sec:2-3}

Bayesian filtering has become a cornerstone technique in many visual tasks, offering a probabilistic approach to understanding and predicting dynamic systems~\cite{wang2023fully,stengaard2019imperfect}. Its evolution from simple Bayesian methods to sophisticated filtering algorithms parallels the advancements in visual processing and computational power. At its core, Bayesian filtering involves updating beliefs over time using Bayes theorem~\cite{puga2015bayes}. Approaches like the Kalman filter~\cite{kalman1960new} and particle filter~\cite{doucet2001introduction} have been pivotal in visual tasks, each with its strengths in handling linear and non-linear systems, respectively~\cite{kutschireiter2017nonlinear}. The intersection of Bayesian filtering with machine learning has opened new frontiers. Bayesian filtering excels in dealing with uncertainty and noise in visual data, which is crucial in tasks like object recognition~\cite{pei2022bayesian}, object detection~\cite{wang2022narrowing}, gesture recognition~\cite{alrowais2023hand}, and scene understanding~\cite{lee2020multimodal}. Recent breakthroughs by Yu \et~\cite{yu2022adaptive}, which introduced a recursive Bayesian filtering technique for remaining useful life estimation of degrading systems, and Su \et~\cite{su2023prognostic}, which showcased improved performance in breast cancer prediction using a hybrid Bayesian network model. We hold the opinion that Bayesian filtering continues to play a pivotal role in advancing visual task processing. Its ability to manage uncertainty and integrate with emerging technologies positions it as a critical player in the future of image and video analysis.

\section{Proposed Method}\label{sec:3}

DBF is the main work in this study. As outlined earlier, prevailing approaches predominantly use infrared information to determine the position of the target object. DBF, however, innovatively incorporates motion data to address the challenges of TIR tracking in complex scenarios. In the DBF framework, motion and infrared information from the target object are sourced from the system and observation models and then integrated for TIR tracking. Section~\ref{sec:3-1} briefly revisits the principle framework of Bayesian filtering. Section~\ref{sec:3-2} is dedicated to explaining how the TIR tracking issue is modeled using Bayesian filtering. The construction of the system model and observation model specific to DBF are detailed in Sections~\ref{sec:3-3} and~\ref{sec:3-4}, respectively.

\subsection{Revisit Bayesian Filtering}\label{sec:3-1}

Bayesian filtering deals with estimating a system's hidden state based on a sequence of observations. In the context of TIR tracking, if we consider the target position as the hidden state and the raw infrared information from each frame as the observation, then TIR tracking essentially becomes a specialized application of Bayesian filtering. However, it should be noted that Bayesian filtering itself only offers a theoretical framework for state estimation and requires the establishment of precise system and observation models for practical application to a specific problem.

The system state and the observations are the most essential concepts in Bayesian filtering, and they have the following important characteristics:
\begin{enumerate}
  \item The system state is the target to be estimated and is not directly observable.
  \item The system state affects the observations that can be observed.
  \item The system state is Markovian with respect to the observations, \ie the system state at the next moment is only related to the state at the current moment and is independent of both the system state and the observations at past moments.
\end{enumerate}

A state space can describe the above system characteristics, which consists of the following two models:

The \textbf{system model} describes the relationship between the system state at adjacent moments $(t-1)$ and $t$, which is a Markov chain according to the Markovianity of the system state and can be expressed as:
\begin{equation}\label{eq:2-1}
  p(s_t|s_{t-1})=p(s_t|s_{t-1},z_{1:t-1})
\end{equation}
where $s_t$ denotes the system state at moment $t$, and $z_{1:t-1}$ denotes the observations from moment 1 to $(t-1)$.

The \textbf{observation model} describes the effect of the system state on the observations. It is the conditional probability distribution $p(z_t|s_t)$ of the observations $z_t$ when the system state is $s_t$.
\begin{figure}[t]
\begin{center}
    \includegraphics[width=\linewidth]{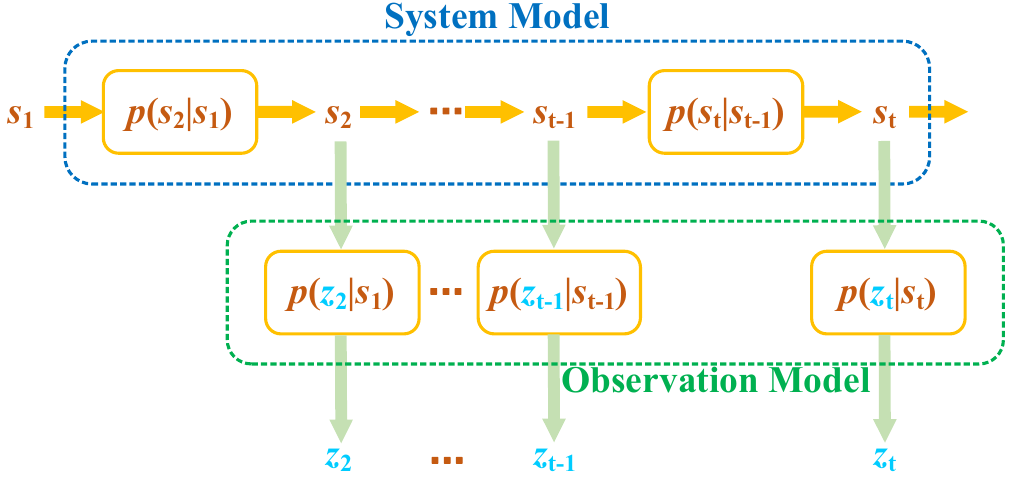}
\end{center}
\vspace{-1.5em}
   \caption{Overview of the state space of DBF. It consists of system model and observation model.}
\label{fig:bayesian}
\end{figure}

Fig.\ref{fig:bayesian} illustrates the relationship between the system state, observations, system model, and observation model. It can be seen that the update of the system state is only related to the system state and the system model at the current moment, while the update of the observations is only associated with the system state and the observable model at the current moment. Over time, the system state changes and produces observations.

The objective of Bayesian filtering is to estimate the system state $s_{1:t}$, given a series of observations $z_{1:t}$, a known system model $p(s_t|s_{t-1})$, and an observation model $p(z_t|s_t)$. This step of the estimation process is iterative, and each step requires an estimate of the current system state $s^*_t$. The objective function can therefore be formulated as follows:
\begin{equation}\label{eq:2-2}
  s_t^*=\arg\max p(s_t|z_{1:t}),\;t>1
\end{equation}
where $s_1$ is the initial state.

The iterative formula between $p(s_{t-1}|z_{1:t-1})$ and $p(s_t|z_{1:t})$ can solve the objective function (Eq.\ref{eq:2-2}). The derivation is as follows:

Based on the total probability theorem, the following equation can be obtained as:
\begin{equation}\label{eq:2-3}
  p(s_t|z_{1:t-1})=\sum_{s_{t-1}}p(s_t|s_{t-1},z_{1:t-1})p(s_{t-1}|z_{1:t-1})
\end{equation}
Combining Eqs.\ref{eq:2-1} and \ref{eq:2-3}, we get:
\begin{equation}\label{eq:2-4}
  p(s_t|z_{1:t-1})=\sum_{s_{t-1}}p(s_t|s_{t-1})p(s_{t-1}|z_{1:t-1})
\end{equation}
According to the Bayes rule, the following equation can be obtained as:
\begin{equation}\label{eq:2-5}
  p(s_t|z_{1:t})=\frac{p(z_t|s_t)p(s_t|z_{1:t-1})}{p(z_t|z_{1:t-1})}
\end{equation}
The iterative relationship can be obtained according to Eqs.\ref{eq:2-4} and \ref{eq:2-5} as:
\begin{equation}\label{eq:2-6}
  p(s_t|z_{1:t})=\frac{p(z_t|s_t)\sum_{s_{t-1}}p(s_t|s_{t-1})p(s_{t-1}|z_{1:t-1})}{p(z_t|z_{1:t-1})}
\end{equation}
The denominator is independent of $s_t$ and can be regarded as a constant here. Therefore, the above equation can be simplified as:
\begin{equation}\label{eq:2-7}
  p(s_t|z_{1:t})=\frac{1}{Z_t}p(z_t|s_t)\sum_{s_{t-1}}p(s_t|s_{t-1})p(s_{t-1}|z_{1:t-1})
\end{equation}

Eq.\ref{eq:2-7} is the iterative formula for $p(s_{t-1}|z_{1:t-1})$ to $p(s_t|z_{1:t})$. Here, the initial value $p(s_1|z_1)$ is given, and it is only necessary to obtain $p(s_t|s_{t-1})$ and $p(z_t|s_t)$ at each step to carry the estimation process forward, which are provided by the system and observation models, respectively. Therefore, solving the state estimation problem depends on the specific settings of the system model $p(s_t|s_{t-1})$ and the observation model $p(z_t|s_t)$.

\subsection{TIR Tracking Formalization}\label{sec:3-2}

To solve the TIR tracking problem using Bayesian filtering, we define the system state $s_t$ at time $t$ as the motion data of the target object, \ie the relative variation of the target center coordinates between $(t-1)$ and $t$. The reasons for this definition are as follows.

\begin{enumerate}
  \item The tracking method is concerned with the change in position due to the motion of the target object between consecutive TIR video frames, not the position itself. The next position can be obtained by adding the change to the current position.
  \item The size of the target object is different in TIR video sequences. Therefore, the same absolute change in position has different meanings for different target sizes. As a highly abstract variable, the system state should not depend on a specific target size, so it is more appropriate to use relative center variations.
\end{enumerate}

In the TIR tracking problem, a rectangular box $G=(x, y; w, h)$ is generally used to annotate the target object in a two-dimensional image, where $(x, y)$ are the coordinates of the top-left corner of the rectangle, and $(w, h)$ is the width and height of the rectangular box, respectively. Therefore, the center coordinates of the rectangular box are calculated as:
\begin{equation}
    \left\{
        \begin{aligned}
            & cx=x+w/2  \\
            & cy=y+h/2
        \end{aligned}
    \right.
\end{equation}

The variation in the center coordinates of the rectangular box between $t-1$ and $t$ (where $t>1$) is as follows,
\begin{equation}
    \left\{
        \begin{aligned}
            &\Delta cx_t^\prime=cx_t-cx_{t-1}=x_t+w_t/2-x_{t-1}+w_{t-1}/2  \\
            &\Delta cy_t^\prime=cy_t-cy_{t-1}=y_t+h_t/2-y_{t-1}+h_{t-1}/2
        \end{aligned}
    \right.
\end{equation}

The relative change corresponding to the variation is as follows,
\begin{equation}
    \left\{
        \begin{aligned}
            &\Delta cx_t=\frac{\Delta cx_t^\prime}{w_{t-1}}=\frac{x_t-x_{t-1}+\frac{1}{2}(w_t-w_{t-1})}{w_{t-1}}  \\
            &\Delta cy_t=\frac{\Delta cy_t^\prime}{h_{t-1}}=\frac{y_t-y_{t-1}+\frac{1}{2}(h_t-h_{t-1})}{h_{t-1}}
        \end{aligned}
    \right.
\end{equation}
where $(\Delta cx_t, \Delta cy_t)$ is the system state $s_t$. During the tracking process, the position of the target object at time $T$ can be expressed as $\sum^T_{t=2}s_t^*$, where $s_t^*$ denotes the best estimate of the system state at time $t$.

For the observation $z_t$ in the Bayesian filtering framework, we define it as the raw infrared image of the $t$-th frame. The observation model estimates the distribution of the system state from the infrared information of the current frame. Thus, the tracking problem can be solved by estimating the optimal system state $\{s_t^*\,|\,t = 1, 2, \ldots, T\}$.

\subsection{System Model}\label{sec:3-3}

The system model is defined as the probability distribution $p(s_t|s_{t-1})$ of the current state $s_t$ under the condition that the system state $s_{t-1}$ at the previous frame. It is used to give an a priori estimate of the likelihood of the current state of the system based on the previous state, \ie the prior distribution, before the observation $z_t$ acts at the current moment.
We established the system model using 2D Brownian motion, which describes the tendency of a target object to move on its own without external influences. It is independent of the observation model and unrelated to infrared information. This situation is similar to the motion of particles in physics when they are not subject to external forces.

For a 2D Brownian motion $\mathcal{B}$, it satisfies the following properties:

\begin{enumerate}
  \item The initial value is 0, \ie $\mathcal{B}_0=(0, 0)$.
  \item The function $t \rightarrow \mathcal{B}_t$ is continuous at $t$ almost everywhere.
  \item The increment of $\mathcal{B}_t$ is independent, and for $t>s$, satisfies $(\mathcal{B}_t-\mathcal{B}_s)$ independently of any $\mathcal{B}_u$, where $0\leq u\leq s$.
  \item The increment $(\mathcal{B}_t-\mathcal{B}_s)$ obeys a 2D Gaussian distribution with mean $(0, 0)$ and covariance $t\mathbf{I}$.
\end{enumerate}

We model the DBF framework with 2D Brownian motion and define $\mathcal{B}_t$ as the target position $(x, y)$ at time $t$. The system state $s$ is the difference $(\Delta x^\prime, \Delta y^\prime)$ between two adjacent frames, which is also the increment of $\mathcal{B}_t$. Their relationship can be expressed as follows,
\begin{equation}\label{eq:3-4}
    s_t=(x_t,y_t)-(x_{t-1},y_{t-1})=\mathcal{B}_t-\mathcal{B}_{t-1},\quad t>0
\end{equation}
where the units of the difference $(\Delta x^\prime, \Delta y^\prime)$ are the scale of the DBF state space, not the pixel distance of the raw infrared image.

Since the initial target position $(x_0, y_0)$ is specified in the first video frame, we set this position as the origin to satisfy Property 1, \ie $\mathcal{B}_0=(0, 0)$. Although the computer calculates and stores the position $(x, y)$ in discrete form, we can assume that the position in DBF state space is continuous to satisfy Property 2. The fact that the pixel distances, in practice, are just discrete samples of the position in DBF space does not affect the effectiveness of our approach.
In Property 3, $\mathcal{B}_t$ has independent increments implies that $(\mathcal{B}_{t_1}-\mathcal{B}_{s_1})$ and $(\mathcal{B}_{t_2}-\mathcal{B}_{s_2})$ are independent if $0\leq s_1 < t_1 \leq s_2 < t_2$ --- the combination of Eq.\ref{eq:3-4} leads to the conclusion that $s_t$ and $s_{t-1}$ are independent. So the system model $p(s_t|s_{t-1})$ can be simplified as $p(s_t)=p(\Delta x^\prime, \Delta y^\prime)$.
According to Property 4, the increment $s_t$ obeys a 2D Gaussian distribution with mean $(0, 0)$ and covariance $t\mathbf{I}$. Since the time interval $t$ is 1 here, the covariance can be expressed as $\mathbf{I}$. It further follows that both $\Delta x^\prime$ and $\Delta y^\prime$ obey a 1D Gaussian distribution $\mathcal{N}(0, 1)$ and are independent of each other.

Based on the above analysis, the system model can be represented as follows,
\begin{equation}
    \left\{
        \begin{aligned}
            & p(s_t|s_{t-1})=p_x(\Delta x^\prime)p_y(\Delta y^\prime)  \\
            & p_x(\Delta x^\prime)=\frac{1}{\sqrt{2\pi}}e^{-{\Delta x^\prime}^2}  \\
            & p_y(\Delta y^\prime)=\frac{1}{\sqrt{2\pi}}e^{-{\Delta y^\prime}^2}
        \end{aligned}
    \right.
\end{equation}

It is worth noting that the scale units in DBF state space are not equivalent to the actual image scale. Therefore, we use the coefficients $\lambda_x$ and $\lambda_y$ to transform the unit scale from the image to the state space. The final system model is as follows,
\begin{equation}
    \left\{
        \begin{aligned}
            & p(s_t|s_{t-1})=p_x(\Delta x)p_y(\Delta y)  \\
            & p_x(\Delta x)=\frac{\lambda_x}{\sqrt{2\pi}}e^{-(\lambda_x\Delta x)^2}  \\
            & p_y(\Delta y)=\frac{\lambda_y}{\sqrt{2\pi}}e^{-(\lambda_y\Delta y)^2}
        \end{aligned}
    \right.
\end{equation}
where $\Delta x$ and $\Delta y$ is the target position difference in the raw image.

\subsection{Observation Model}\label{sec:3-4}

The observation model is used to evaluate the probability $p(z_t|s_t)$ that the most recent observation $z_t$ occurs in the current system state $s_t$. It obtains a likelihood estimate of the system state after obtaining the observations. It is used to make a likelihood estimation of the system state after obtaining the observations, correcting the probability distribution predicted by the system model.

The observation model in this study is based on the detection methods. This is due to the many similarities between the two approaches as follows,
\begin{enumerate}
  \item Each candidate corresponds to each possible system state $s_t^i$.
  \item The current frame image is the observation $z_t$.
  \item The detection method generates a response score based on the feature similarity between the template and the candidate window in the current frame, and the more similar to the template, the higher the response score. The response scores are highly consistent with the role of the observation model output $p(z_t|s_t)$.
\end{enumerate}

\begin{figure}[t]
\begin{center}
    \includegraphics[width=\linewidth]{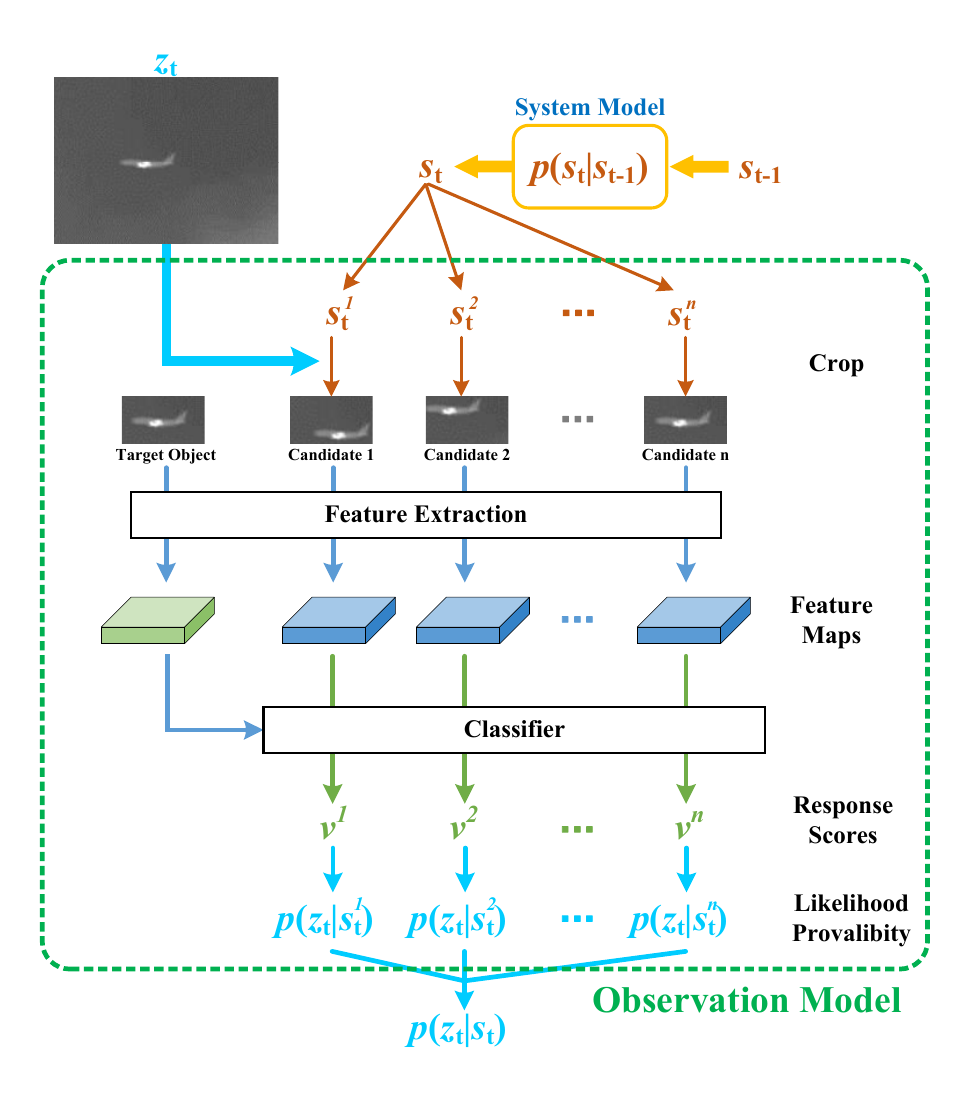}
\end{center}
\vspace{-1.5em}
\caption{Overview of the Observation Model. The observation $z_t$ is the $t$-th frame raw image. Candidates are cropped on $z_t$ according to the system state generated by the system model, and the feature maps of each candidate are obtained. $v_i$ is the response score output by the classifier to the system state $s_t^i$ based on the candidate region. The classifier is learned through the feature maps of the target object.}
\label{fig:model}
\end{figure}

Based on this analysis, we propose an observation model based on the tracking-by-detection method. As shown in Fig.\ref{fig:model}, the pipeline of the observation model mainly consists of two modules: feature extraction and classifier.

\begin{table}[t]\footnotesize
\centering
\caption{Illustration of the feature extraction network.}
\label{tab:vgg}
\begin{tabular}{ccc}
\toprule
Layers      & Configuration & Strides \\
\midrule
Convolution & 96$\times$ 7$\times$ 7      & 2       \\
Activation  & ReLU+LRN      & -       \\
Pooling     & Maxpooling    & 2       \\
Convolution & 256$\times$ 5$\times$ 5       & 2       \\
Activation  & ReLU+LRN      & -       \\
Pooling     & Maxpooling    & 2       \\
Convolution & 512$\times$ 3$\times$ 3       & 1       \\
Activation  & ReLU          & -\\
\bottomrule
\end{tabular}
\end{table}

\begin{table}[t]\footnotesize
\centering
\caption{Illustration of the classification network.}
\label{tab:cls}
\begin{tabular}{ccc}
\toprule
Layers      & Configuration \\
\midrule
Convolution & 512$\times$ 1$\times$ 1       \\
Activation  & ReLU+LRN             \\
Convolution & 512$\times$ 1$\times$ 1            \\
Activation  & ReLU+LRN           \\
Convolution & 2$\times$ 1$\times$ 1           \\
Loss  & Softmax         \\
\bottomrule
\end{tabular}
\end{table}

We choose VGG-M~\cite{vgg} as the network architecture for feature extraction. We use the activation value of the third convolutional layer as the deep feature for image representation, and all other parameters are kept consistent with the original VGG-M architecture, as shown in Table~\ref{tab:vgg}.
Inspired by MDNet~\cite{mdnet}, we use a neural network with three fully connected layers as the classifier. The tracking task involves only two categories, \ie foreground (target object) and background, instead of the 1000 categories of the ImageNet classification task~\cite{imagenet}. Moreover, the former uses low-level convolutional features, unlike the latter, which uses high-level convolutional features~\cite{lcsn}. Therefore, we reduce the number of units in the intermediate hidden layer from 4,096 to 512 dimensions for the general classification task. This also reduces the costs of computation and storage. The specific network parameters are shown in Table \ref{tab:cls}, where the fully connected layer is actually implemented with a $1\times 1$ convolutional layer, and the last layer is a softmax function that outputs a 2D vector $\mathbf{v}=(v_1, v_2)$, where $v_1$ and $v_2$ denote the probability that the candidate is foreground and background, respectively. Since $v_1+v_2=1$, we take $v_1$ as the response score.

The final output of the observation model is the likelihood probability $p(z_t|s^i_t)$, so it is also necessary to transform the response score into a probabilistic form. We use a weighted softmax function to perform the following transformation as follows,
\begin{equation}
  p(z_t|s^i_t)=\frac{e^{\alpha^iv^i}}{\sum_je^{\alpha^jv^j}}
\end{equation}
where $v_i$ is the response score of the system state $s^i_t$. It can be seen that the higher the response score is, the higher the likelihood probability is. $\alpha^i$ is a penalty hyperparameter that obeys the Gaussian distribution and controls the impact of the response score on the probability distribution, \ie the closer $s_t^i$ is to the center of the previous frame, the larger $\alpha^i$ is.

\section{Experiments}\label{sec:4}

In this section, we detail the experimental results that highlight the need and effectiveness of applying and revitalizing Bayesian filtering for TIR object tracking. Section \ref{sec:4-1} outlines the implementation details of our DBF. Section \ref{sec:4-2} describes the benchmark datasets used and the metrics for evaluation. The ablation studies are presented in Section \ref{sec:4-3}, followed by the comparison results with state-of-the-arts on the LSOTB-TIR~\cite{lsotb} and PTB-TIR~\cite{ptb} benchmark datasets in Section \ref{sec:4-4}. Finally, Section \ref{sec:4-5} offers more discussions to improve the performance of DBF.

\subsection{Implementation Details}\label{sec:4-1}

Our TIR object tracking method, DBF, was developed using the MatConvNet~\cite{vedaldi2015matconvnet} toolbox and trained on both RGB datasets such as COCO~\cite{coco}, LaSOT~\cite{lasot}, and GOT-10k~\cite{got10k}, and the TIR dataset LSOTB-TIR~\cite{lsotb}. All the experiments were conducted on a cloud server with an Intel$^\circledR$ Xeon$^\circledR$ E5-2680 v4 CPU @ 2.4GHz CPU with 128GB RAM, and a NVIDIA$^\circledR$ GeForce RTX$^{\mathrm{TM}}$ 2080 Ti GPU with 11GB VRAM. We trained the observation model using Stochastic Gradient Descent (SGD) with a batch size of 8 and a momentum of 0.9. As described in Section \ref{sec:3-4}, we used a modified VGG-M~\cite{vgg} as the base feature extractor. The classifier's settings adhered to the default configurations of MDNet~\cite{mdnet}. The initial training phase of the observation model spanned 60 epochs on the RGB datasets, with the learning rate exponentially decayed from $10^{-2}$ to $10^{-5}$. This was followed by a fine-tuning phase of 30 epochs on the TIR dataset, setting the learning rate at $10^{-3}$ and the weight decay at $10^{-5}$.

\subsection{Datasets and Evaluation Metrics}\label{sec:4-2}

\textbf{Datasets:} PTB-TIR~\cite{ptb} dataset comprises 60 manually annotated infrared pedestrian sequences, amassing more than 30,000 frames in total. Each sequence has nine attribute labels for the attribute-based evaluation. LSOTB-TIR~\cite{lsotb} dataset is notably larger, offering 1400 infrared video sequences with high-quality annotations, totaling in excess of 600,000 frames. It features five varieties of moving objects -- people, animals, vehicles, aircraft, and boats -- in four distinct scenarios and is evaluated on 12 challenging attributes, thereby making it the most extensive TIR dataset to date.

\textbf{Evaluation metrics:} Three distinct metrics are employed in LSOTB-TIR~\cite{lsotb} dataset to evaluate the performance of TIR trackers. Precision is defined as the ratio of frames in which the Center Location Error (CLE) falls below a specified threshold (20 pixels), relative to the total number of frames. Normalized Precision adjusts for the resolution and size of the predicted bounding box, mitigating the impact of varying target sizes. The effectiveness of TIR trackers is assessed using the Area Under Curve (AUC) of Normalized Precision within the range of 0 to 0.5. Success is measured by the proportion of frames where the overlap ratio (OR) between the predicted and ground truth bounding boxes exceeds a predefined threshold. For the PTB-TIR~\cite{ptb} dataset, precision and success are used to assess the performance of the TIR trackers.

\subsection{Ablation Studies}\label{sec:4-3}

In this section, we first analyze the proposed DBF method on two benchmark datasets to evaluate the effectiveness of the feature extractor network architecture and training dataset size.

\begin{table*}[t]\footnotesize
\centering
\caption{Comparison of the proposed DBF refers to the network architectures of the feature extractor.}
\label{tab:net}
\begin{tabular}{cccccccc}
\toprule
\multirow{2}{*}{Network} &  & \multicolumn{3}{c}{LSOTB-TIR}        &  & \multicolumn{2}{c}{PTB-TIR} \\ \cline{3-5} \cline{7-8}
                         &  & Success & Precision & Norm. Precision &  & Success     & Precision     \\
\midrule
AlexNet                  &  & 0.603        & 0.752          & 0.692               &  & 0.607            & 0.825              \\
VGG-M                    &  & 0.625        & 0.770          & 0.703               &  & 0.626            & 0.839              \\
VGG-16                   &  & 0.617        & 0.760          & 0.697               &  & 0.613            & 0.831              \\
\bottomrule
\end{tabular}
\end{table*}

\textbf{Network architecture:} Among existing TIR object tracking methods, the commonly used CNN architectures are AlexNet~\cite{alexnet}, VGG-M~\cite{vgg}, and VGG-16~\cite{vgg}. Table \ref{tab:net} presents the results of the proposed DBF using these different architectures as the feature extractor. Notably, VGG-M surpasses AlexNet and VGG-16, registering the best scores across all evaluation metrics on both benchmark datasets. AlexNet, an earlier architecture, is smaller in size and offers faster computation but lacks in performance. It is typically employed in early tracking methods where speed is prioritized. VGG-M, in comparison, is wider than AlexNet. The initial two convolutional layers of AlexNet use $5\times 5$ and $3\times 3$ kernels, whereas VGG-M uses $7\times 7$ and $5\times 5$ kernels in its first two layers. Additionally, VGG-M boasts more convolutional channels in layers 2, 3, and 4 than AlexNet and incorporates batch normalization to enhance training efficiency. In practice, many TIR tracking methods utilize only the shallow features of VGG-M, which allows for rapid computation while delivering robust performance, making it a popular choice among convolutional feature extractors. The characteristic of VGG-16 is that it is relatively deep, and replaces a single layer of large convolution kernels with multiple layers of $3\times 3$ small convolution kernels. VGG-16, known for its depth and use of smaller $3\times 3$ convolution kernels, does not perform as well in tracking tasks as it does in classification. This could be due to tracking tasks relying more on shallow features.

\begin{table*}[t]\footnotesize
\centering
\caption{Comparative results of different models using different training dataset aggregation strategies.}
\label{tab:dataset}
\begin{tabular}{lccccccc}
\toprule
\multirow{2}{*}{Training Dataset} &  & \multicolumn{3}{c}{LSOTB-TIR}        &  & \multicolumn{2}{c}{PTB-TIR} \\ \cline{3-5} \cline{7-8}
                         &  & Success & Precision & Norm. Precision &  & Success     & Precision     \\
\midrule
Only-RGB                 &  & 0.596   & 0.749     & 0.680          &  & 0.606       & 0.795         \\
Only-TIR                 &  & 0.603   & 0.742     & 0.679          &  & 0.597       & 0.781         \\
All                      &  & 0.625   & 0.770     & 0.703          &  & 0.626       & 0.839         \\
\bottomrule
\end{tabular}
\end{table*}

\textbf{Training dataset size:} To train the observation model of DBF, a mix of RBG and TIR datasets was used. The RGB datasets encompassed COCO~\cite{coco}, LaSOT~\cite{lasot}, and GOT-10k~\cite{got10k}. COCO is a comprehensive dataset designed for image detection and semantic segmentation, containing over 330,000 images and 1.5 million objects. LaSOT, a newer benchmark for long-term visual tracking, includes 1,400 video sequences, each averaging 2,500 frames, spanning 70 class categories with 16 sequences per category in its training set. GOT-10k, another challenging benchmark, consists of 10,000 video sequences for training purposes. The training subset of the LSOTB-TIR~\cite{lsotb} was also utilized. Table \ref{tab:dataset} illustrates the effectiveness of DBF using different combinations of training datasets. The combined use of RGB and TIR datasets achieves better results than training solely with either RGB or TIR datasets. This indicates that the integration of RGB and TIR datasets exploits the unique properties of RGB and TIR images, thereby enhancing the robustness of feature extraction for TIR object tracking.

\subsection{Comparison with state-of-the-arts}\label{sec:4-4}

To evaluate our proposed DBF, we compared it with state-of-the-art TIR tracking methods, including CFNet~\cite{cfnet}, DSiam~\cite{dsiam}, ECO-deep~\cite{eco}, ECO-HC~\cite{eco}, ECO-STIR~\cite{ecostir}, HSSNet~\cite{hssnet}, MDNet~\cite{mdnet}, MLSSNet~\cite{mlssnet}, MMNet~\cite{mmnet}, SiamFC~\cite{siamfc}, SiamTri~\cite{siamtri}, SRDCF~\cite{srdcf}, TADT~\cite{tadt}, VITAL~\cite{vital}, and UDT~\cite{udt}.

\begin{figure*}[htb]
\begin{center}
    \subfigure{\includegraphics[width=.32\linewidth]{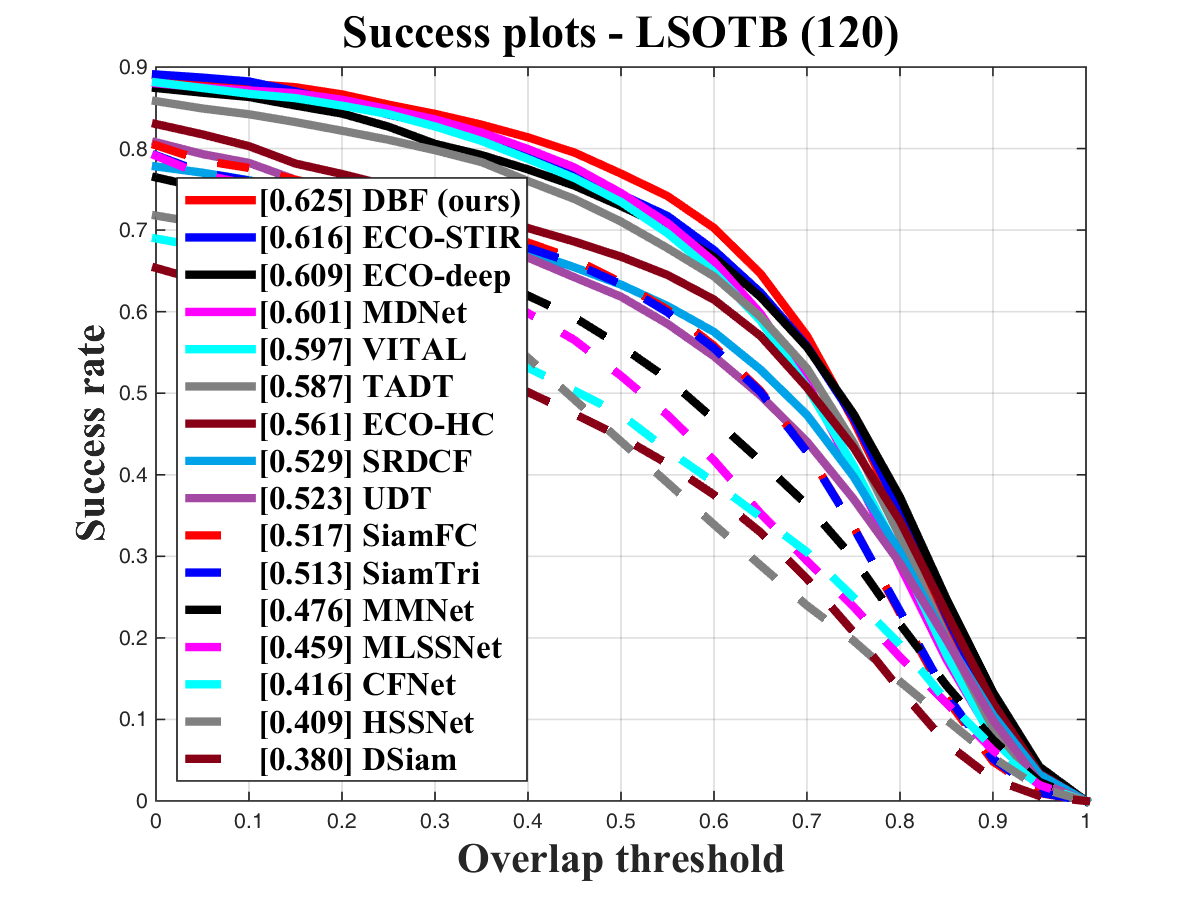}}
    \hspace{0.05em}
    \subfigure{\includegraphics[width=.32\linewidth]{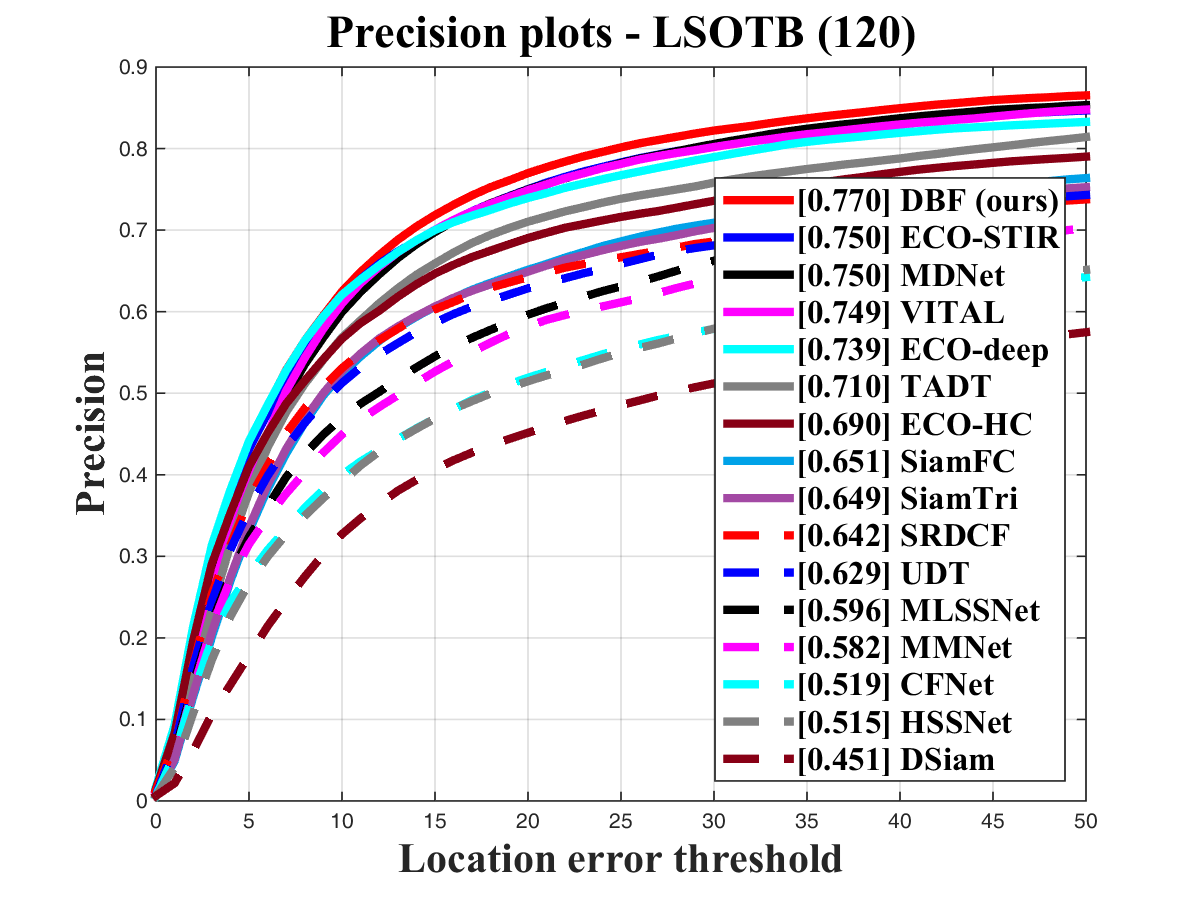}}
    \subfigure{\includegraphics[width=.32\linewidth]{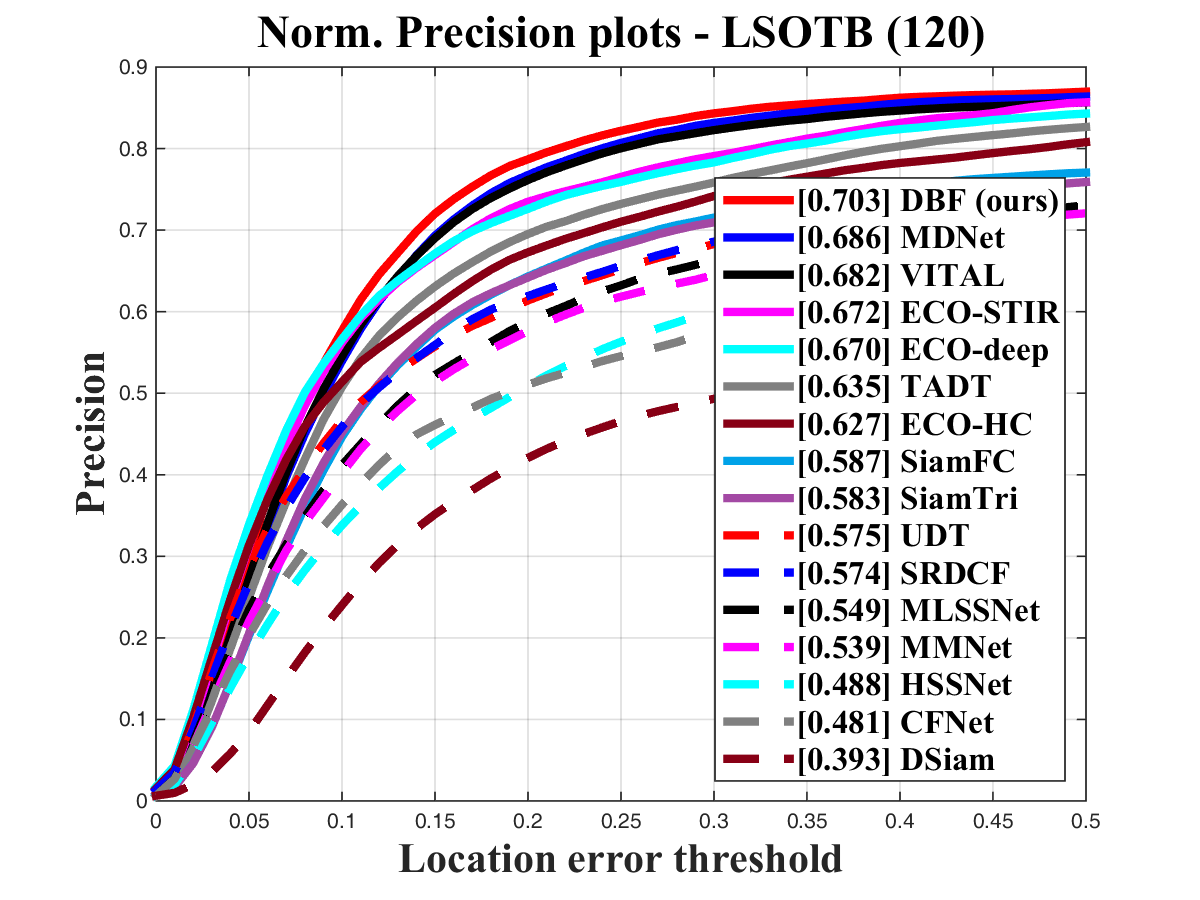}}
\end{center}
\vspace{-0.8em}
\caption{Comparison on the LSOTB-TIR~\cite{lsotb} benchmark dataset.}
\label{fig:lsotb}
\end{figure*}

\subsubsection{Results on LSOTB-TIR}

Fig.\ref{fig:lsotb} shows that DBF achieves top scores in all evaluation criteria, recording 0.625 in success, 0.770 in precision, and 0.703 in normalized precision, outperforming the baseline MDNet~\cite{mdnet} by 2.4\%, 2.0\%, and 1.7\%, respectively. This underscores the effectiveness of Bayesian filtering in learning classifiers for TIR tracking. ECO-deep~\cite{eco}, utilizing factorized continuous convolution operators for multi-resolution deep feature map integration, scores 0.609 in success, 0.739 in precision, and 0.670 in normalized precision. ECO-STIR~\cite{ecostir}, which employs additional synthetic TIR-based deep features for training, records the second-best success and precision scores of 0.616 and 0.750, and the fourth-best normalized precision score of 0.672. This surpasses ECO-deep by 0.7\%, 1.1\%, and 0.2\% in the respective metrics, highlighting the superiority of TIR-based over RGB-based deep features in TIR tracking. Using the LSOTB-TIR training subset, DBF significantly outperforms ECO-STIR with absolute gains of 0.9\%, 2.0\%, and 3.2\% in success, precision, and normalized precision. Notably, classification-based trackers such as VITAL~\cite{vital}, UDT~\cite{udt}, and SRDCF~\cite{srdcf}, with precision scores of 0.749, 0.629, and 0.642, lag behind DBF by more than 2.1\%, 14.1\%, and 12.8\%, respectively. MMNet~\cite{mmnet}, a matching-based tracker with a dual-level feature model, still lags behind DBF by considerable margins of 14.9\%, 18.8\%, and 16.4\% in success, precision, and normalized precision. We attribute the robust performance of DBF to inheriting the strong discriminative capacity of the MDNet classifier. Moreover, DBF outperforms other matching-based trackers such as TADT~\cite{tadt}, SiamFC~\cite{siamfc}, SiamTri~\cite{siamtri}, MLSSNet~\cite{mlssnet}, CFNet~\cite{cfnet}, HSSNet~\cite{hssnet}, and DSiam~\cite{dsiam}, with absolute success score gains of 3.8\%, 10.8\%, 11.2\%, 16.6\%, 20.9\%, 21.6\%, and 24.5\%, respectively. Overall, the results on the LSOTB-TIR benchmark dataset illustrate that the proposed DBF tracker performs favorably against state-of-the-art TIR trackers over the 120 challenging video sequences.

\begin{figure*}[htbp]
\begin{center}
    \subfigure{\includegraphics[width=.32\linewidth]{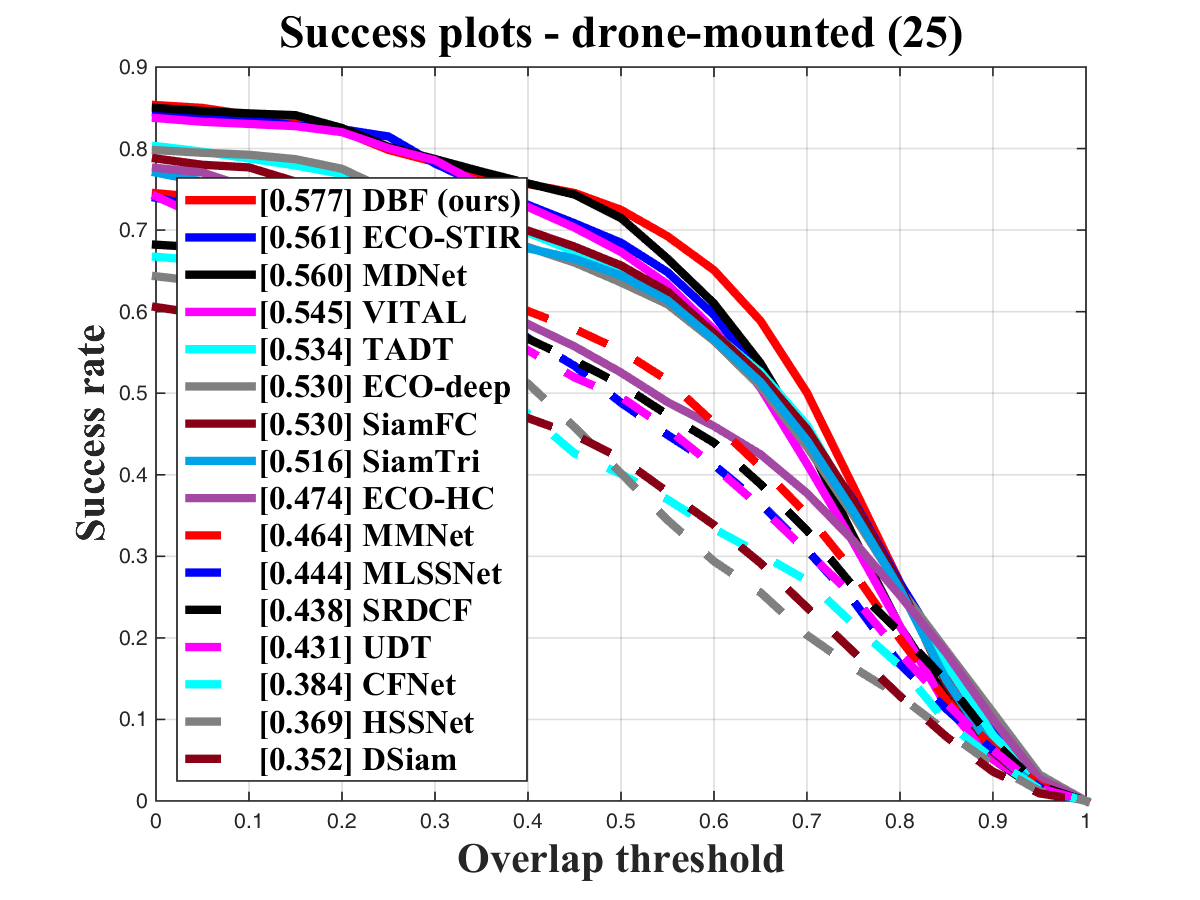}}
    \hspace{0.05em}
    \subfigure{\includegraphics[width=.32\linewidth]{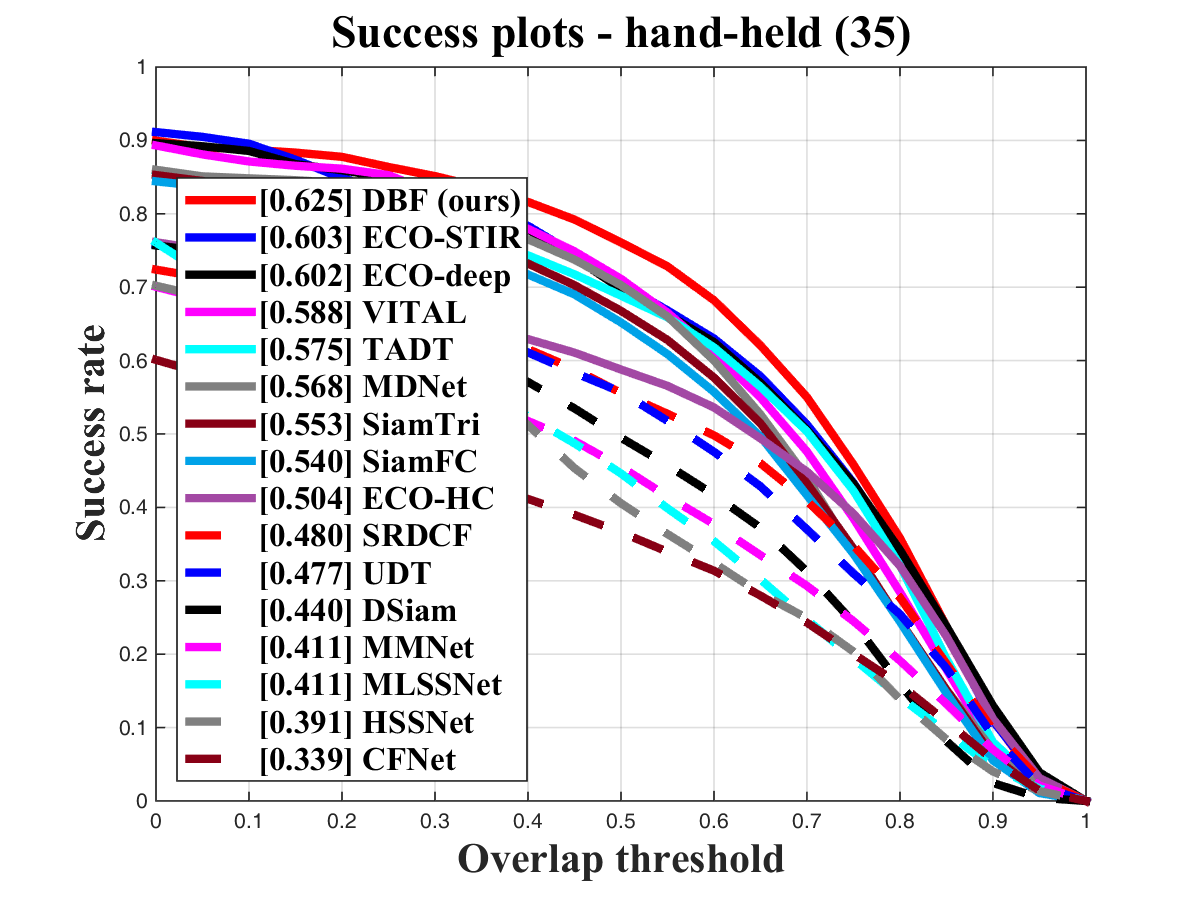}}
    \hspace{0.05em}
    \vfill
    \subfigure{\includegraphics[width=.32\linewidth]{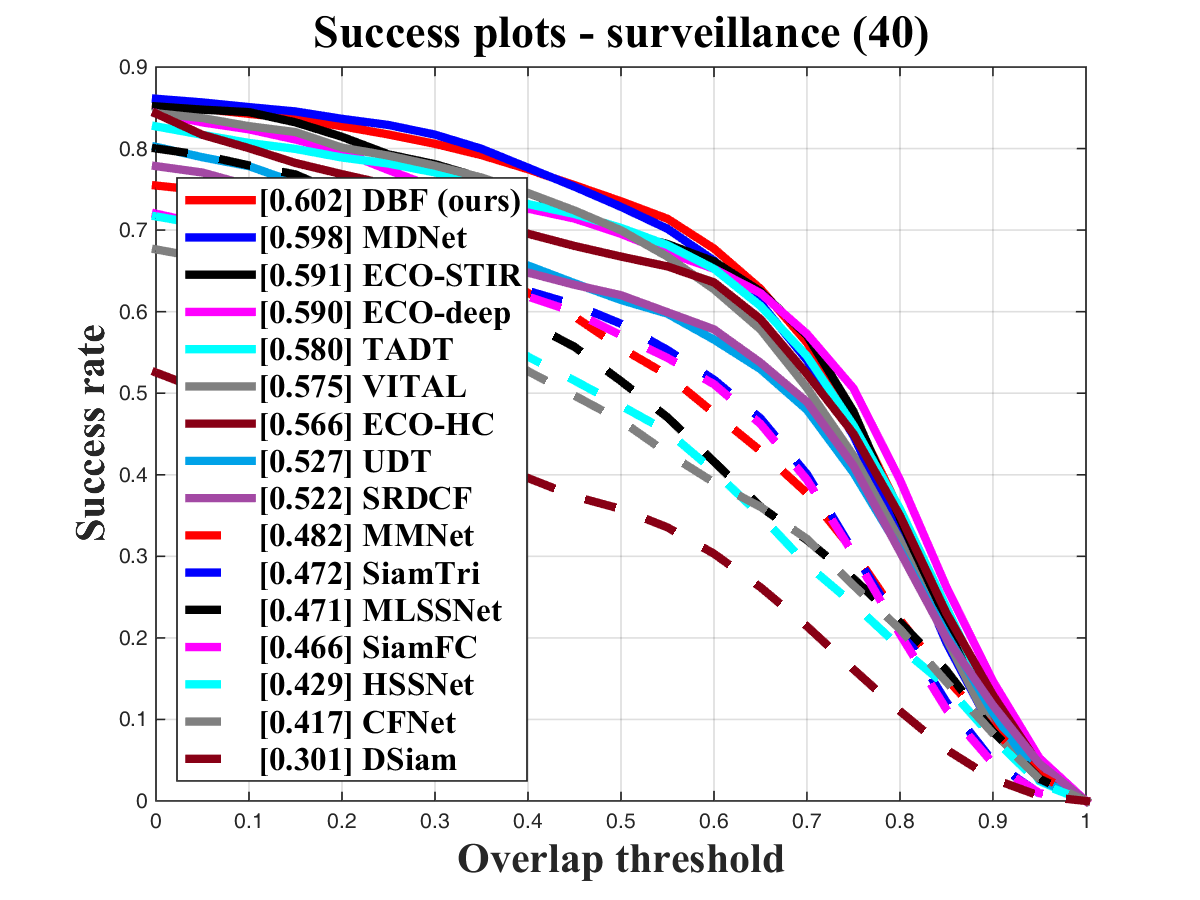}}
    \hspace{0.05em}
    \subfigure{\includegraphics[width=.32\linewidth]{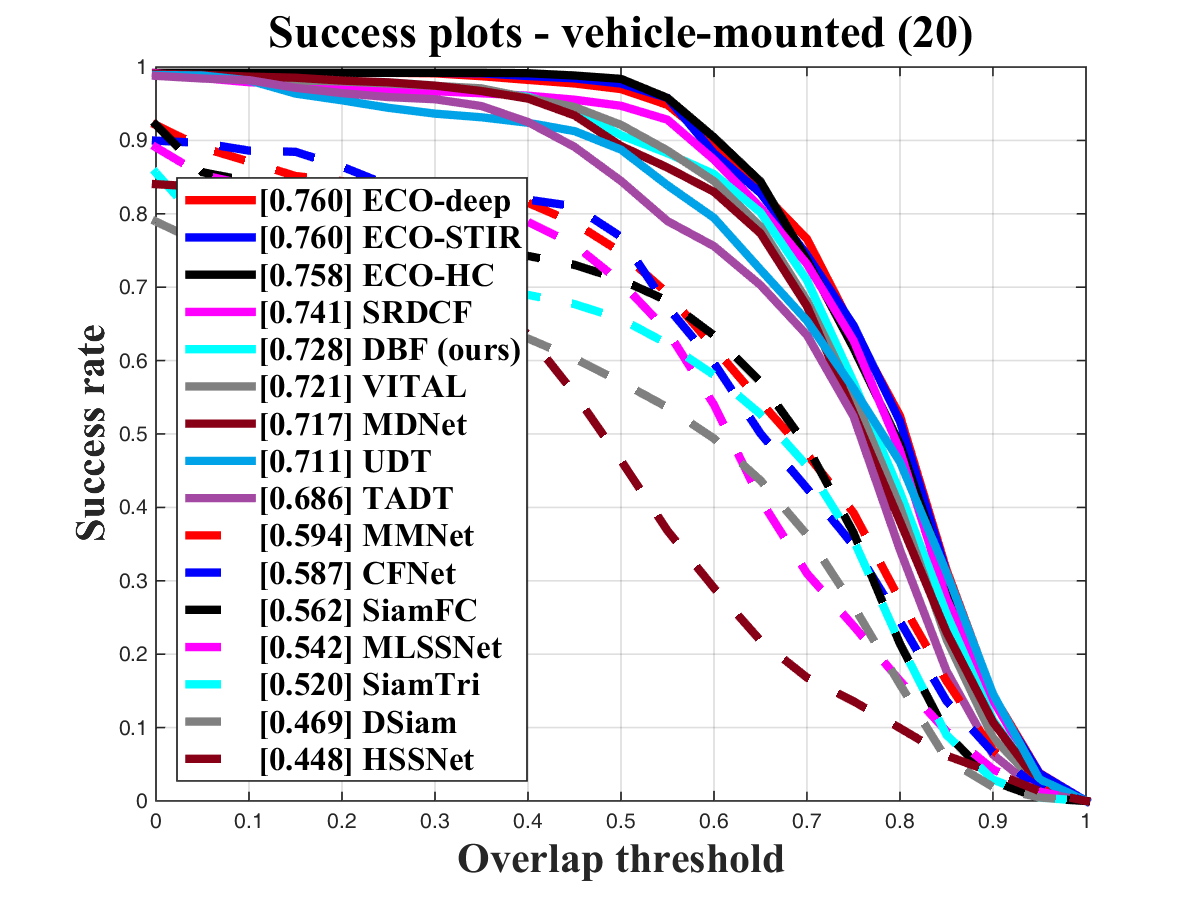}}
\end{center}
\vspace{-0.8em}
\caption{Evaluation of four scenario attribute subsets of the LSOTB-TIR~\cite{lsotb} benchmark dataset.}
\label{fig:lsotbsc}
\end{figure*}

\begin{figure*}[htbp]
\begin{center}
    \subfigure{\includegraphics[width=.32\linewidth]{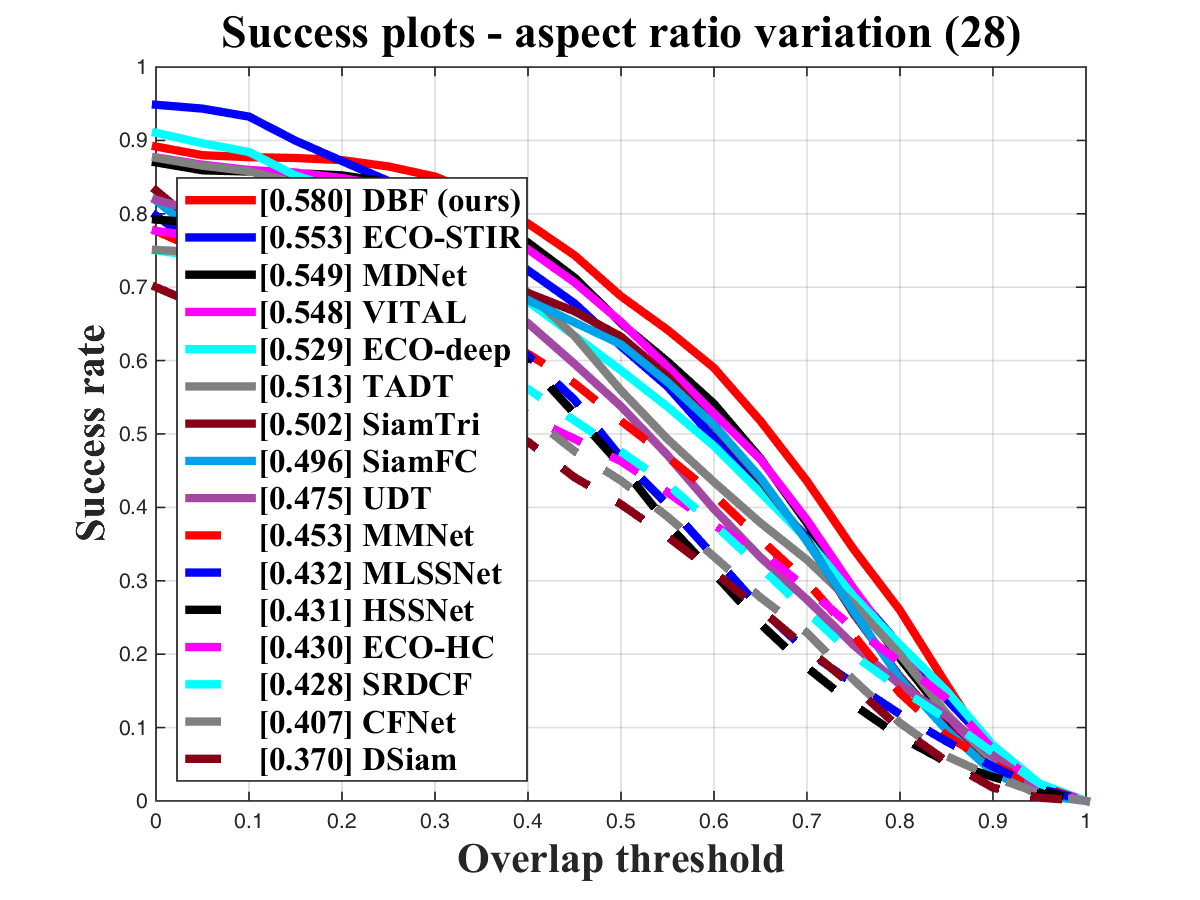}}
    \hspace{0.05em}
    \subfigure{\includegraphics[width=.32\linewidth]{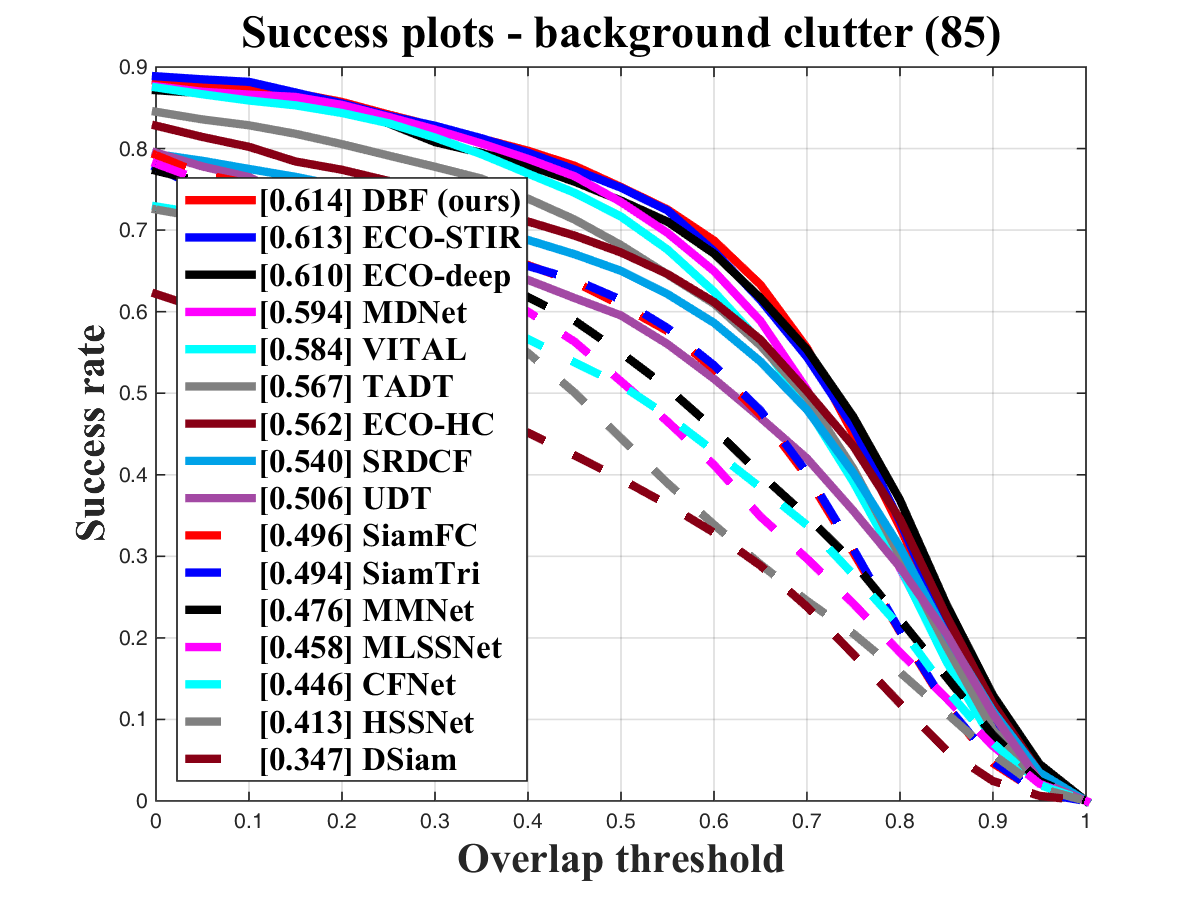}}
    \hspace{0.05em}
    \subfigure{\includegraphics[width=.32\linewidth]{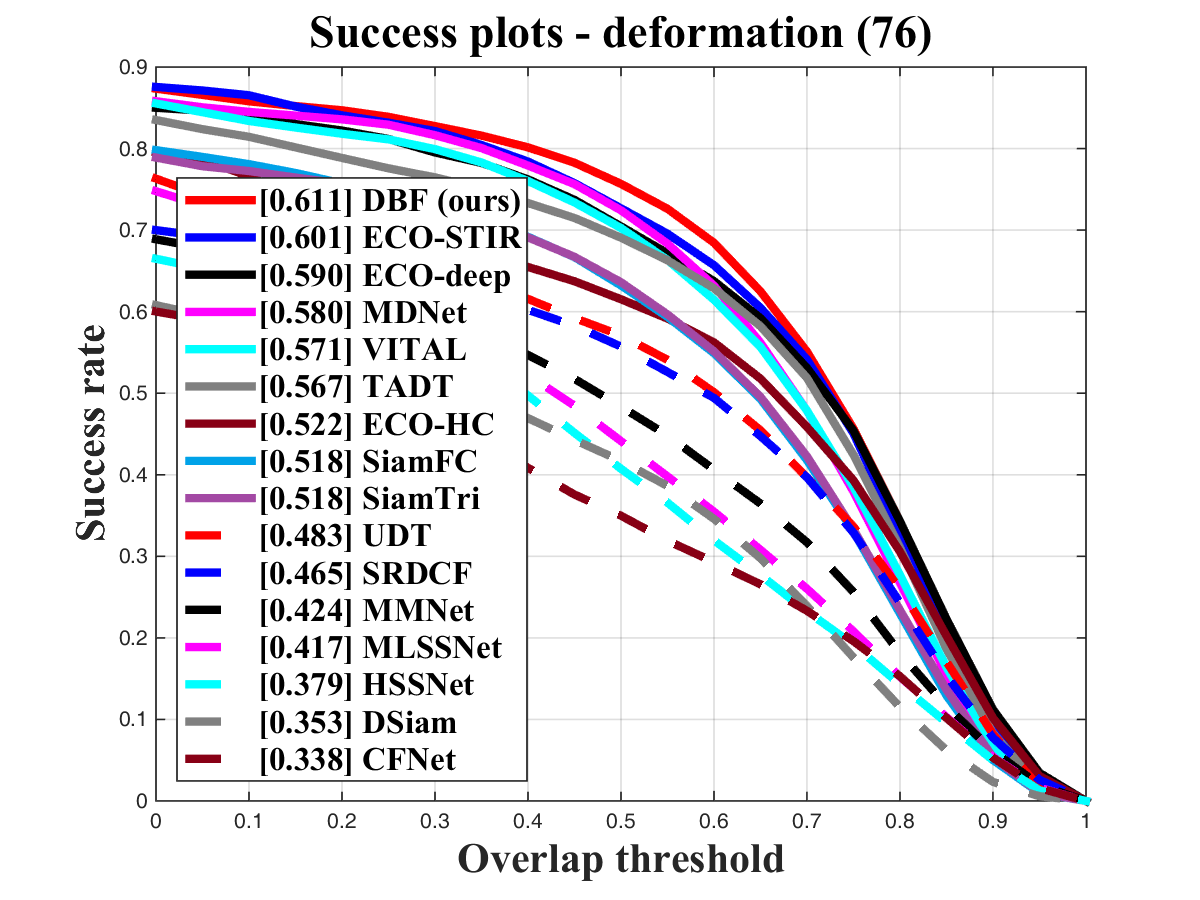}}
    \hspace{0.05em}
    \vfill
    \subfigure{\includegraphics[width=.32\linewidth]{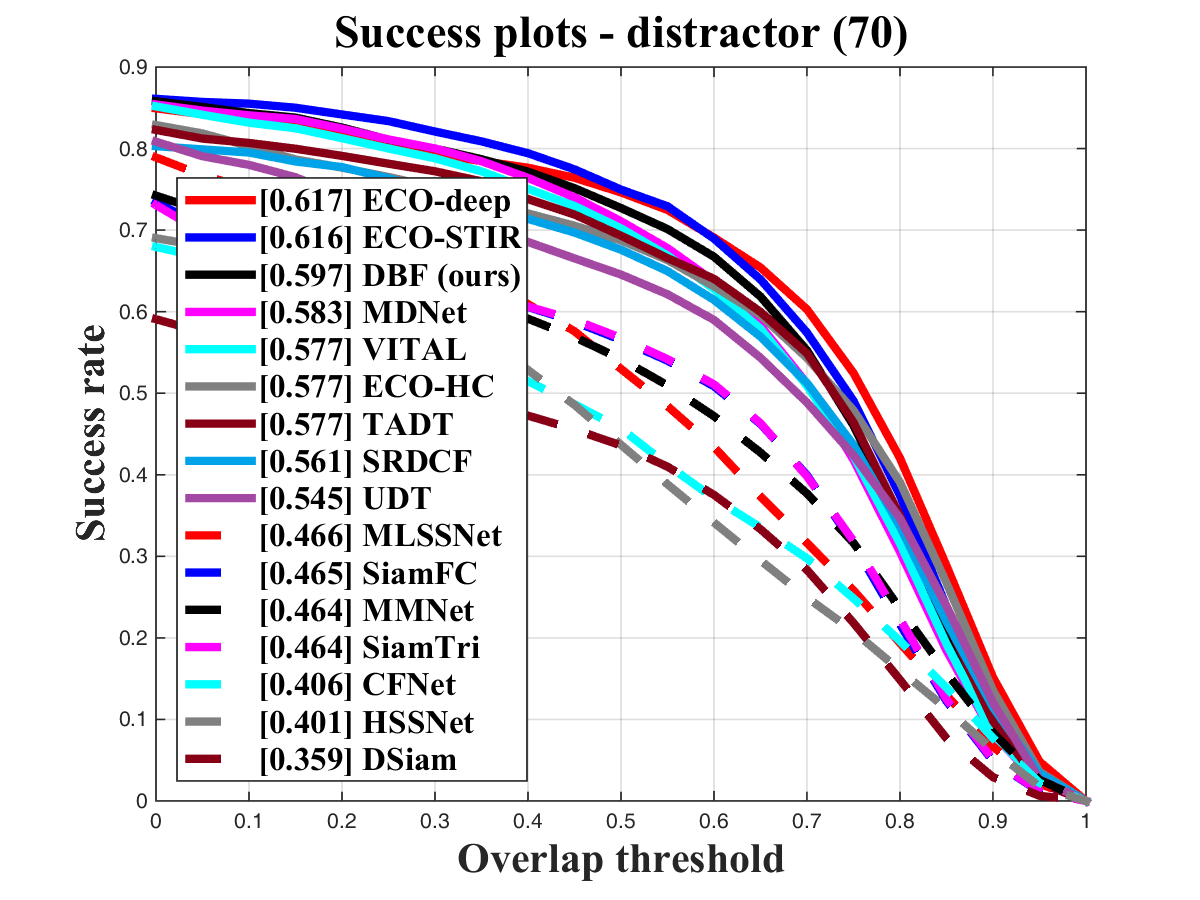}}
    \hspace{0.05em}
    \subfigure{\includegraphics[width=.32\linewidth]{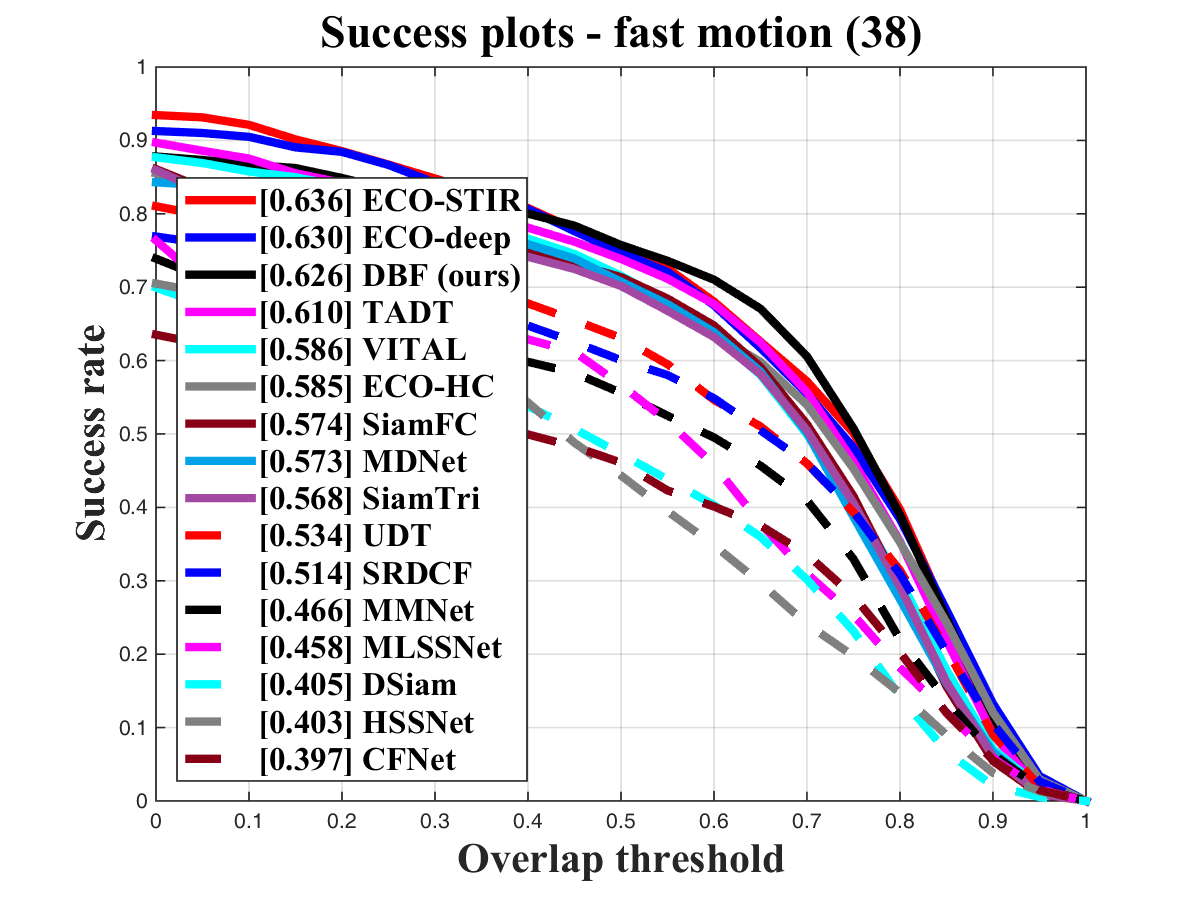}}
    \hspace{0.05em}
    \subfigure{\includegraphics[width=.32\linewidth]{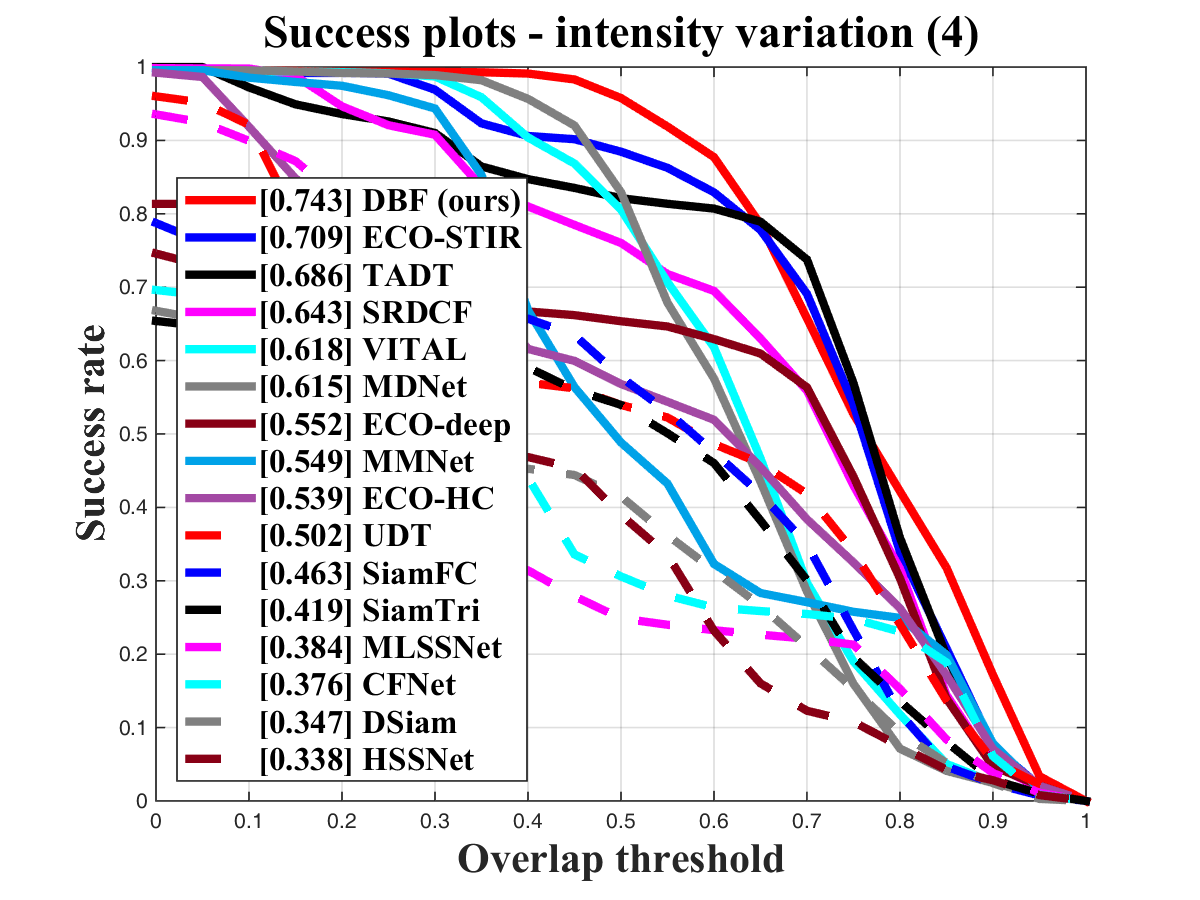}}
    \hspace{0.05em}
    \vfill
    \subfigure{\includegraphics[width=.32\linewidth]{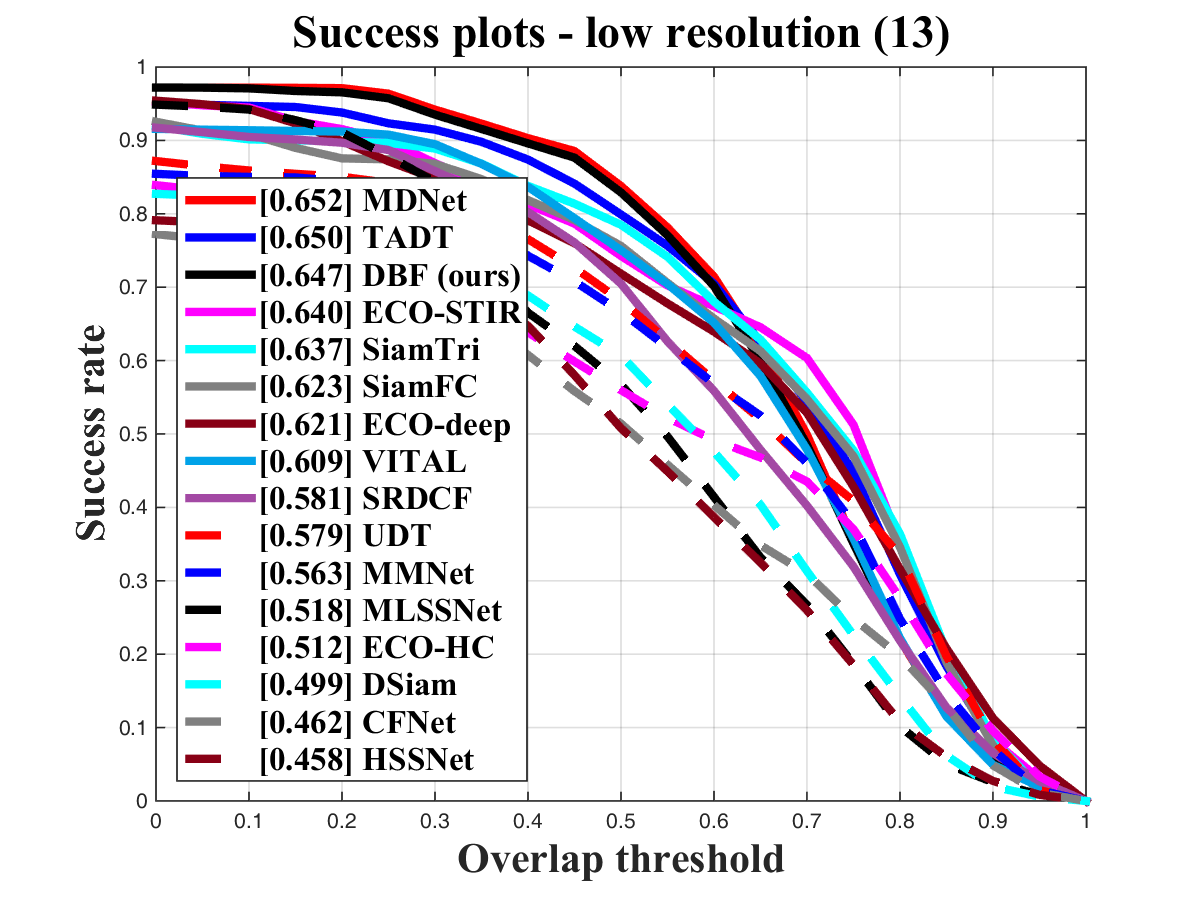}}
    \hspace{0.05em}
    \subfigure{\includegraphics[width=.32\linewidth]{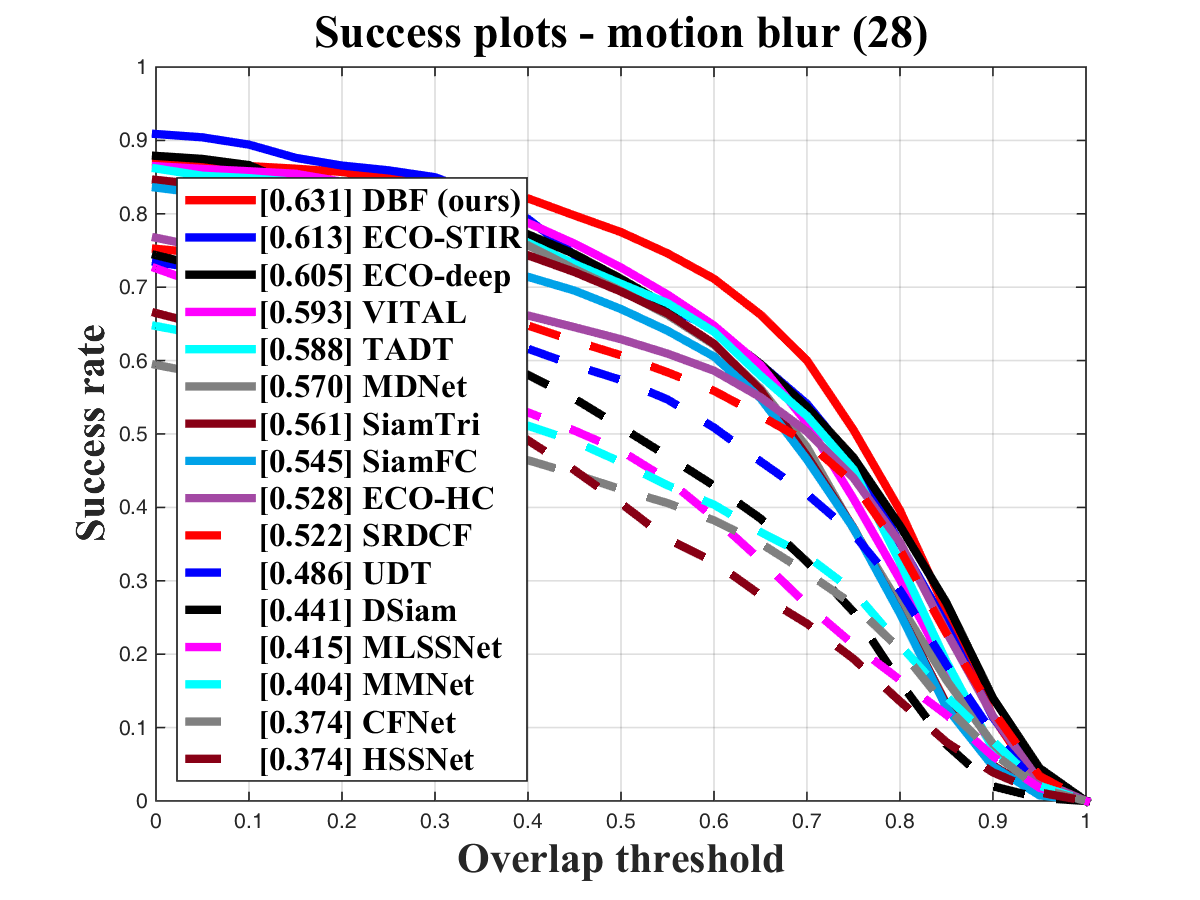}}
    \hspace{0.05em}
    \subfigure{\includegraphics[width=.32\linewidth]{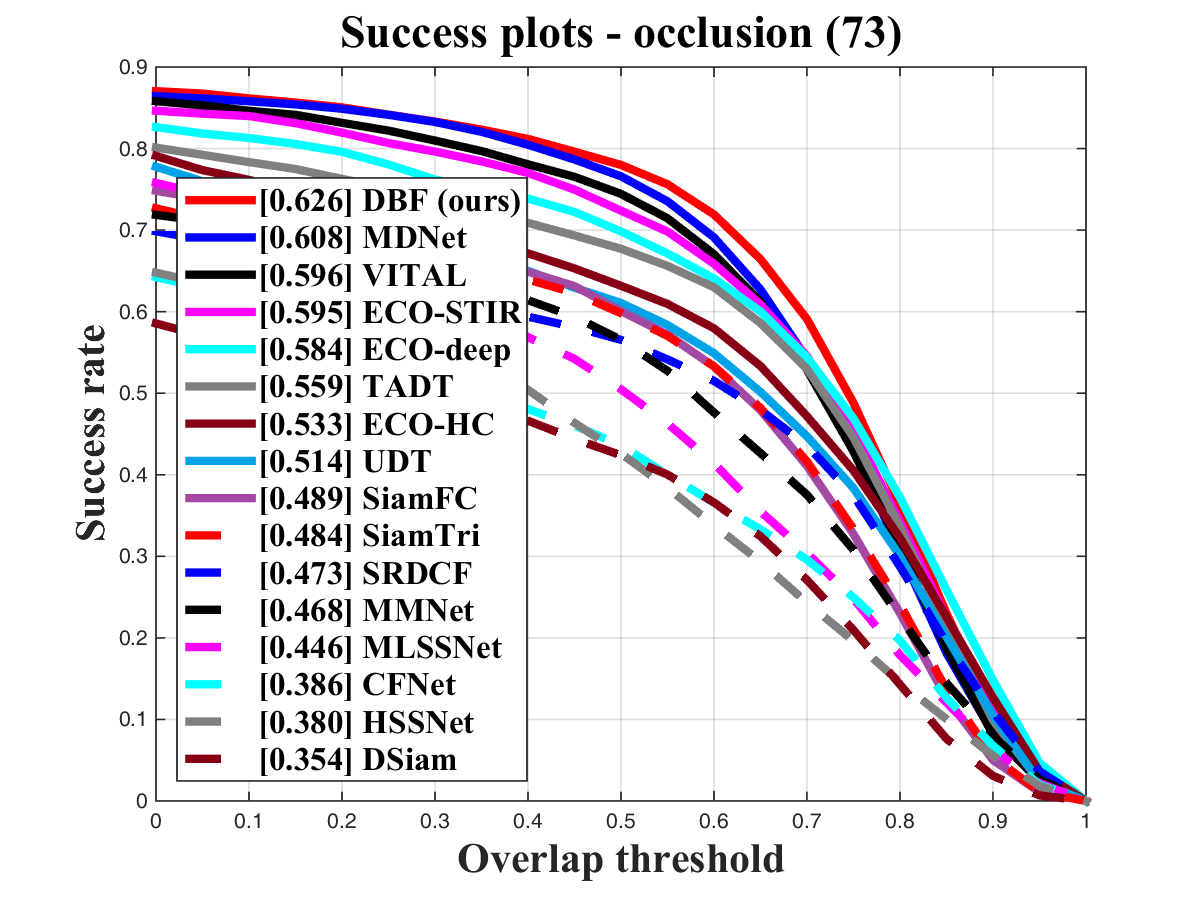}}
    \hspace{0.05em}
    \vfill
    \subfigure{\includegraphics[width=.32\linewidth]{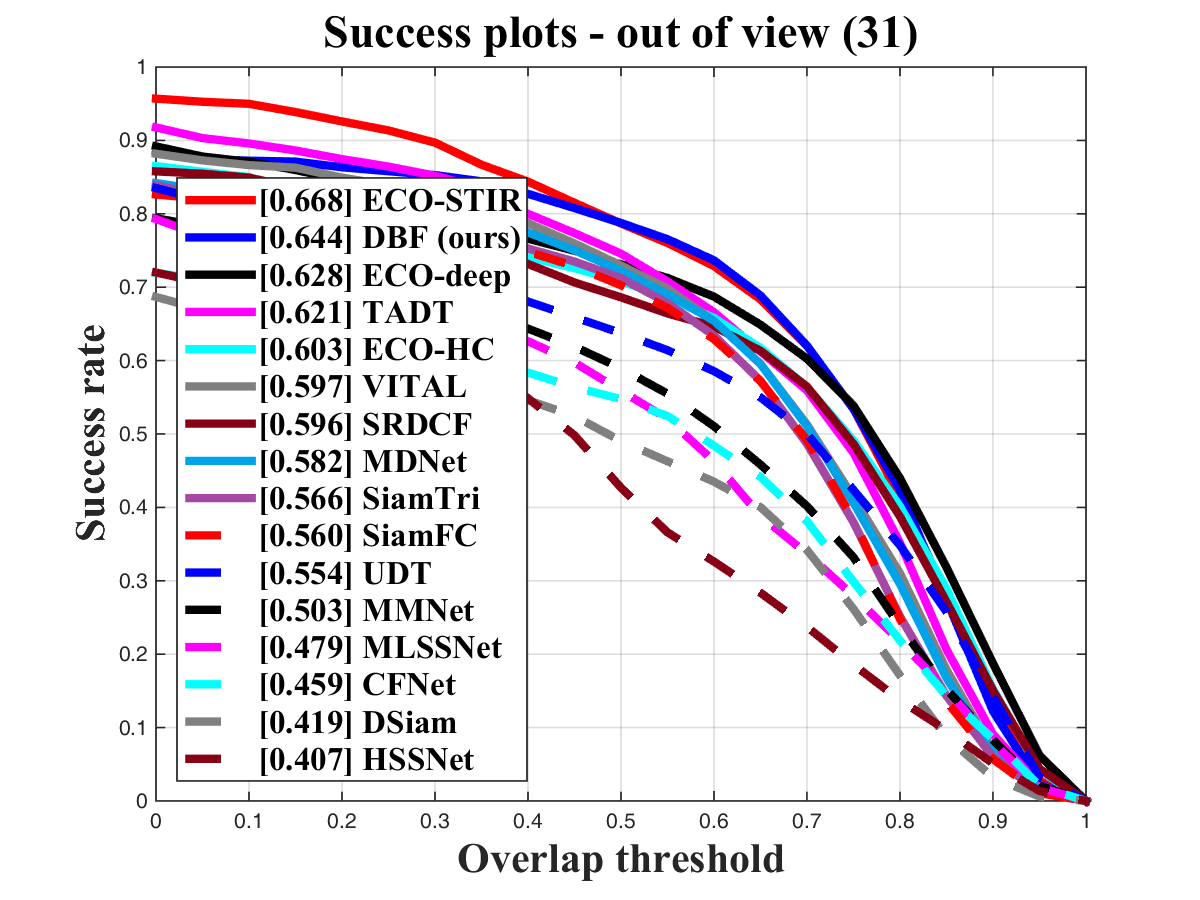}}
    \hspace{0.05em}
    \subfigure{\includegraphics[width=.32\linewidth]{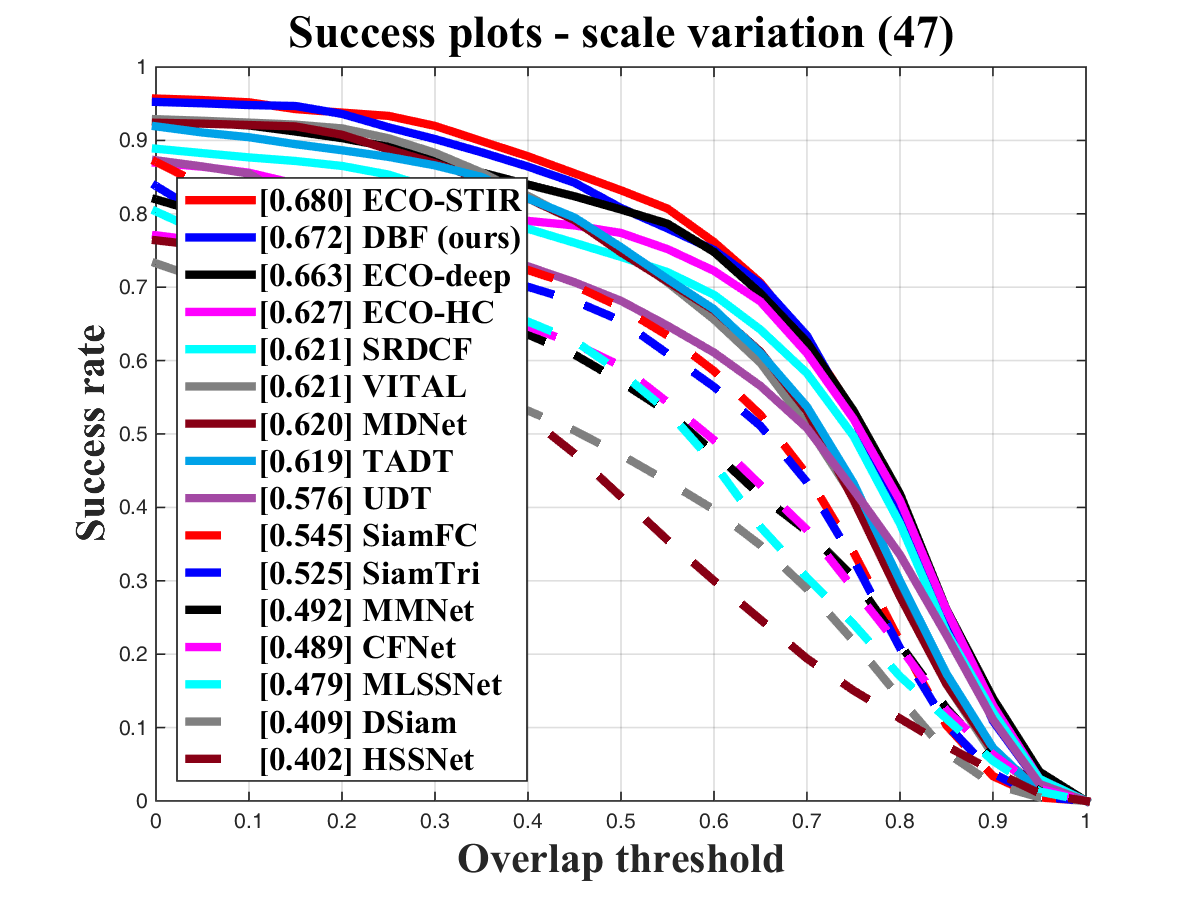}}
    \hspace{0.05em}
    \subfigure{\includegraphics[width=.32\linewidth]{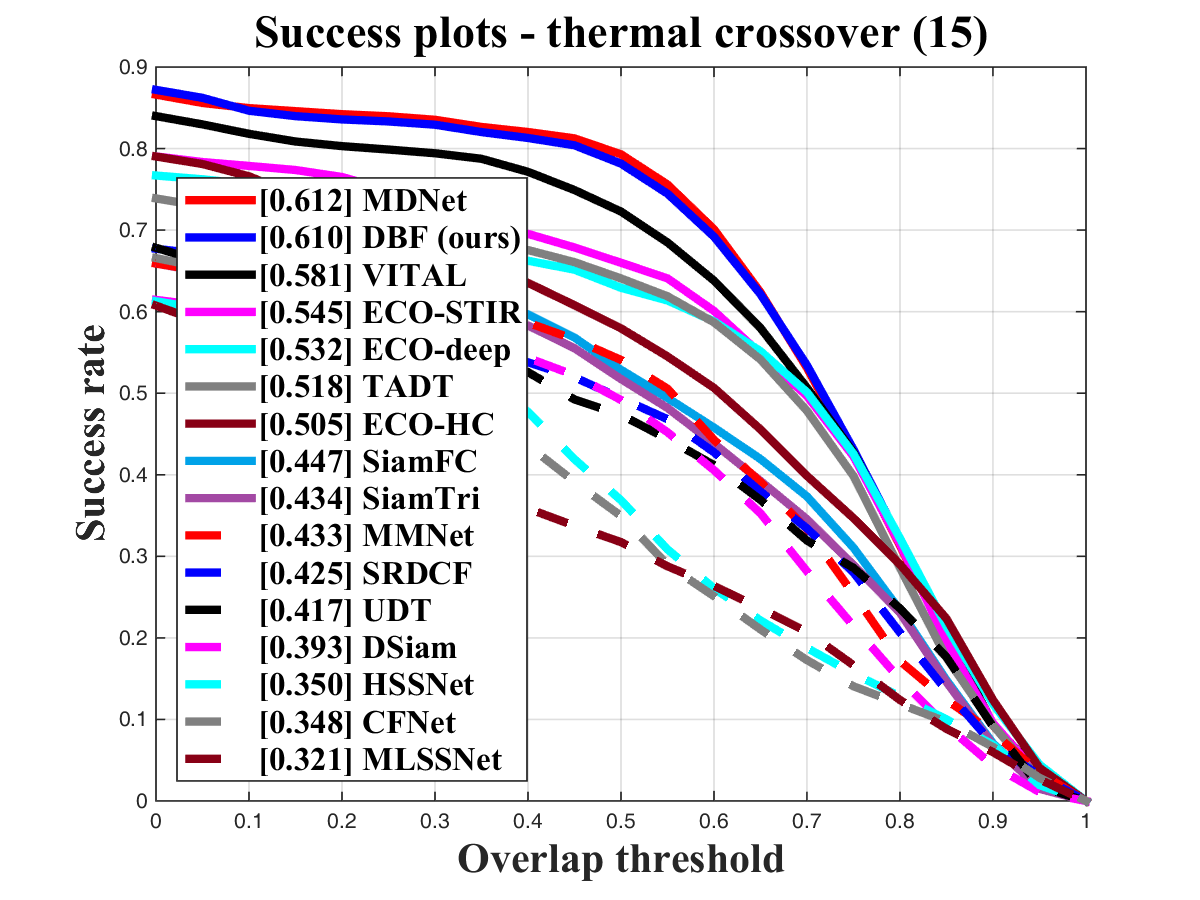}}
\end{center}
\vspace{-0.8em}
\caption{Evaluation of 12 challenge attribute subsets of the LSOTB-TIR~\cite{lsotb} benchmark dataset.}
\label{fig:lsotbatt}
\end{figure*}

In assessing the efficacy of our DBF tracker for various TIR tracking challenges, we benchmarked it against top-tier methods across different attributes. All the 120 video sequences in LSOTB-TIR are divided into four scenarios according to the TIR camera platform: drone-mounted (25 sequences), hand-held (35 sequences), surveillance (40 sequences), and vehicle-mounted (20 sequences). The results of this scenario-based comparison, depicted in Fig.\ref{fig:lsotbsc}, reveal that while DBF underperforms in the vehicle-mounted scenarios, it excels in the other three. Most classification-based methods, including ours, demonstrate top-tier performance due to their effective use of binary classifiers trained on positive and negative samples of the target objects.
Each sequence is also annotated with 12 attributes, posing unique challenges: aspect ratio variation (ARV), BC, DEF, distractor (DIS), fast motion (FM), intensity variation (IV), low resolution (LR), MB, occlusion (OCC), OV, SV, and TC. As Fig.\ref{fig:lsotbatt} shows, DBF demonstrates impressive results across these attributes, achieving the highest success rate in six of them. Although it ranks second in OV, SV, and TC and third in DIS, FM, and LR, its success scores are on par with the top-ranked trackers for each attribute. Specifically, DBF outperforms the base tracker MDNet~\cite{mdnet} across various attributes, indicating the effectiveness of the Bayesian filtering framework in TIR tracking. In addition, DBF outperforms ECO-STIR~\cite{ecostir}, the leading tracker on FM, OV, and SV, with improvements of 2.7\%, 1.0\%, 3.4\%, 1.8\%, 3.1\%, and 6.5\% on ARV, DEF, IV, MB, OCC, and TC, respectively. This comparable performance of DBF can be attributed to the application of Bayesian filtering, which provides enriched motion data and learns the classifier dynamically.

\begin{figure*}[htb]
\begin{center}
    \subfigure{\includegraphics[width=.32\linewidth]{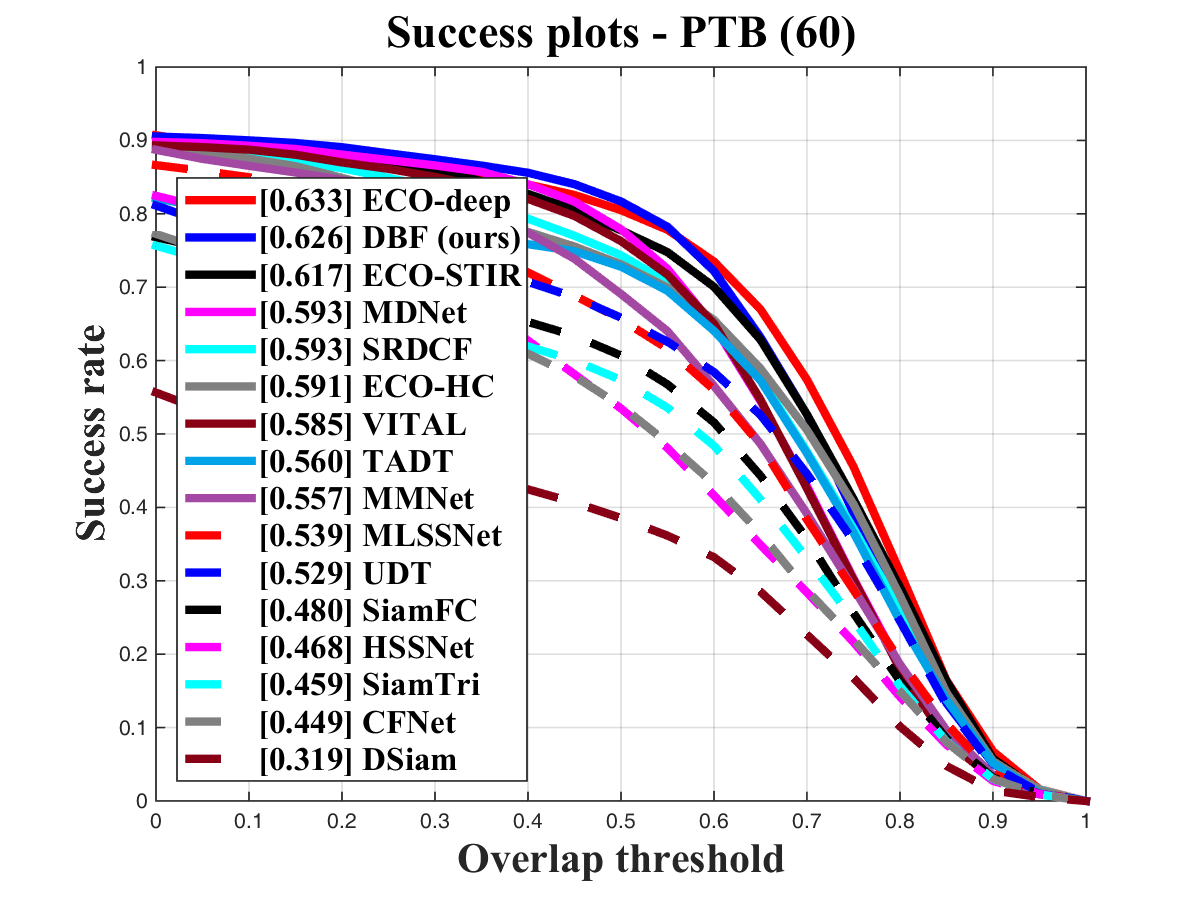}}
    \hspace{0.05em}
    \subfigure{\includegraphics[width=.32\linewidth]{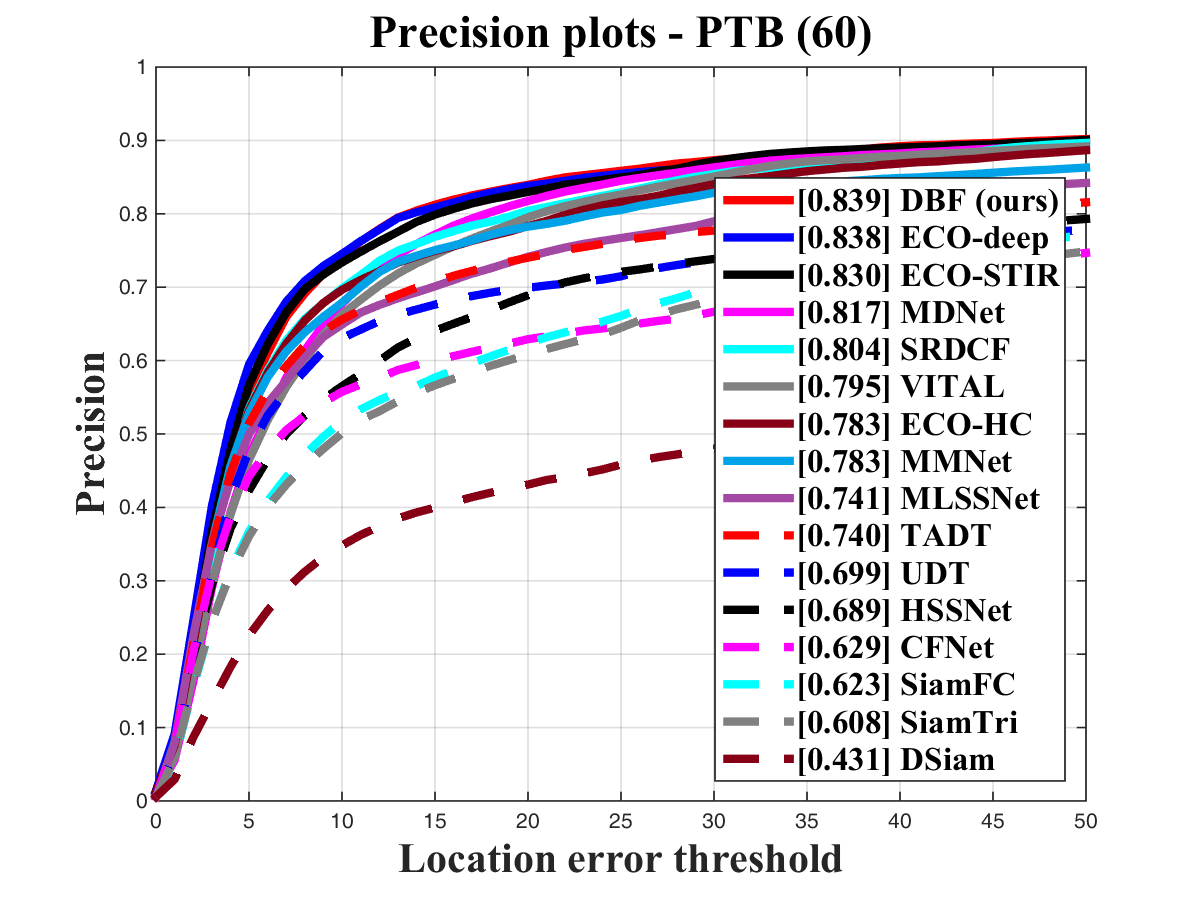}}
\end{center}
\vspace{-0.8em}
\caption{Comparison on the PTB-TIR~\cite{ptb} benchmark dataset.}
\label{fig:ptb}
\end{figure*}

\subsubsection{Results on PTB-TIR}

\begin{figure*}[t]
\begin{center}
    \includegraphics[width=\linewidth]{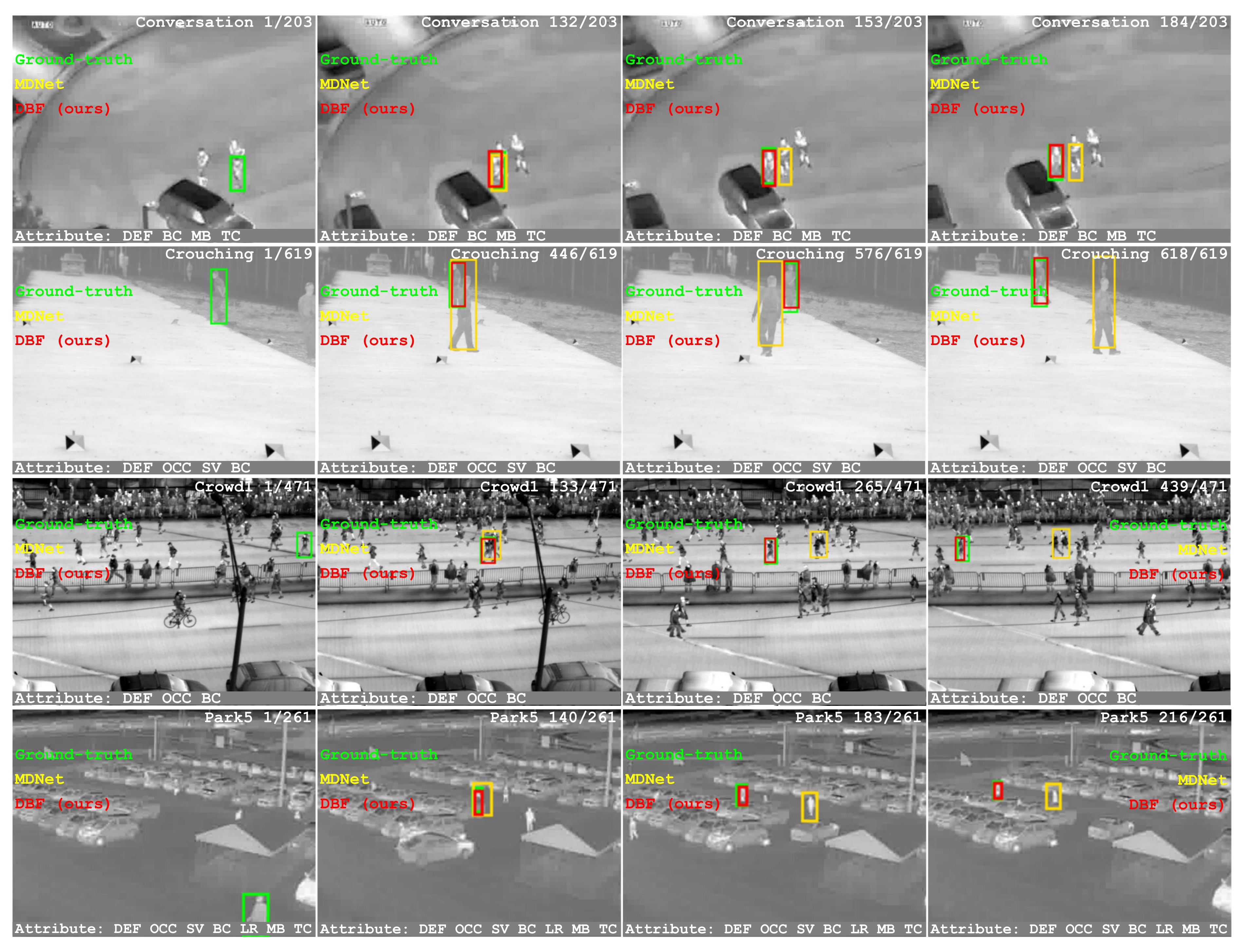}
\end{center}
   \caption{Comparison of our proposed DBF with the baseline MDNet on four challenging sequences (from top to bottom are \emph{conversation}, \emph{crouching}, \emph{crowd1}, and \emph{park5}, respectively).}
\label{fig:visual}
\end{figure*}

\begin{figure*}[t]
\begin{center}
    \includegraphics[width=\linewidth]{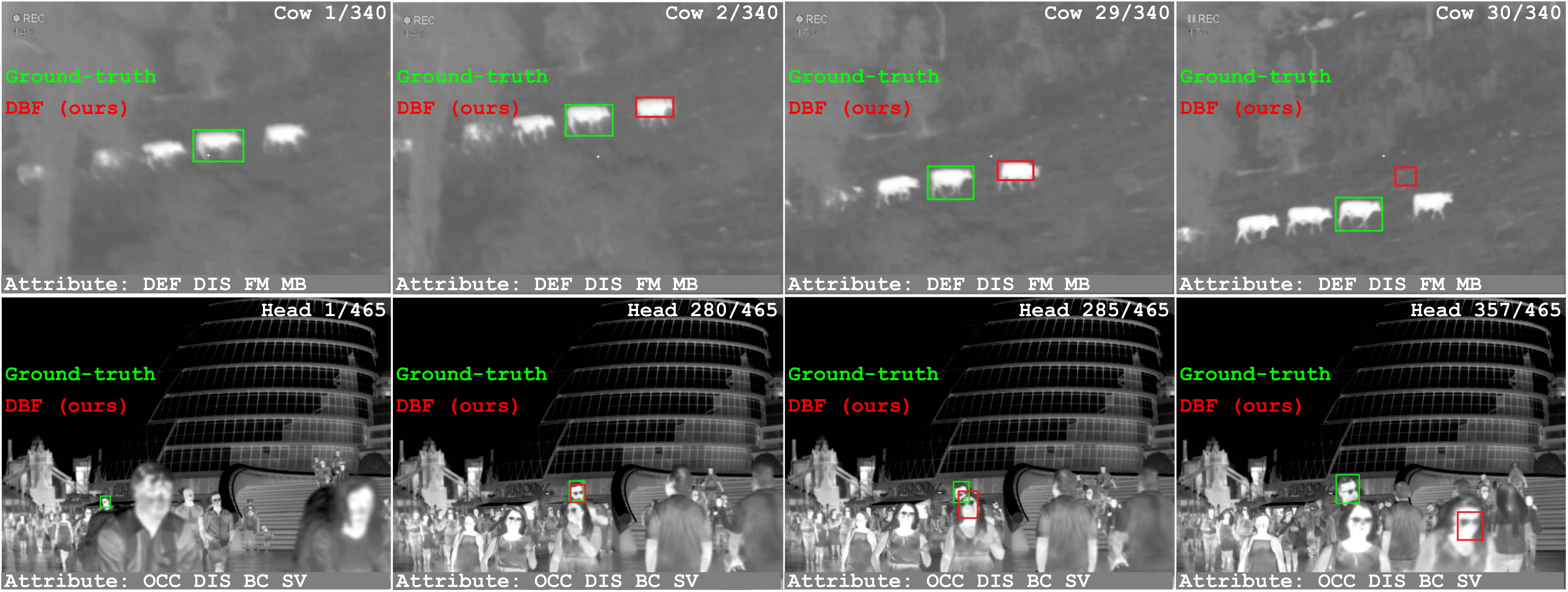}
\end{center}
   \caption{Failure cases on two challenging sequences (from top to bottom are \emph{cow} and \emph{head}, respectively).}
\label{fig:failure}
\end{figure*}

Fig.\ref{fig:ptb} reveals that DBF significantly outperforms the baseline MDNet~\cite{mdnet}, recording a 4.0\% increase in success score and a 2.2\% increase in precision score. This underlines the substantial improvement Bayesian filtering brings to TIR trackers, particularly in pedestrian tracking. DBF attains the highest precision score (0.839), outperforming ECO-deep~\cite{eco}, though ECO-deep slightly leads in success score (0.633). Notably, DBF significantly outperforms ECO-STIR~\cite{ecostir} on the PTB-TIR dataset, despite ECO-STIR being a close competitor on the LSOTB dataset. MLSSNet~\cite{mlssnet}, utilizing a multi-level similarity model for TIR tracking, records a success score of 0.539 and a precision score of 0.741. In comparison, DBF exceeds MLSSNet by 7.6\% and 9.8\% in these metrics, respectively. Additionally, DBF outshines the classification-based tracker VITAL~\cite{vital}, gaining 4.8\% in success and 4.4\% in precision. Furthermore, DBF registers substantial success/precision gains of 3.3\%/3.5\%, 3.5\%/5.6\%, 6.6\%/9.9\%, 6.9\%/5.6\%, 9.7\%/14.0\%, 14.6\%/21.6\%, 15.8\%/15.0\%, 16.7\%/23.1\%, 17.7\%/21.0\%, and 30.7\%/40.8\% compared to SRDCF~\cite{srdcf}, ECO-HC~\cite{eco}, TADT~\cite{tadt}, MMNet~\cite{mmnet}, UDT~\cite{udt}, SiamFC~\cite{siamfc}, HSSNet~\cite{hssnet}, SiamTri~\cite{siamtri}, CFNet~\cite{cfnet}, and DSiam~\cite{dsiam}, respectively. These results highlight DBF's superior and harmonious tracking performance, excelling in both accuracy and robustness.

\subsubsection{Visualized results}

As shown in Fig.\ref{fig:visual}, the visualized results of DBF on four video sequences selected from the PTB-TIR benchmark dataset --- \emph{conversation}, \emph{crouching}, \emph{crowd1}, and \emph{park5} --- are presented. These sequences encompass four challenging attributes: BC, MB, SV, and TC. We chose MDNet~\cite{mdnet} as the baseline for comparative experiments, as our observation model is an extension of it. In BC cases, where the background shares infrared and texture characteristics with the target object, MDNet often misidentifies the background as the target object, leading to tracking inaccuracies. DBF, however, shows improved tracking by leveraging not only infrared information but also an independent system model to estimate the possible position of the target object, thereby reducing background interference. For MB, SV, and TC cases, involving rapid changes in appearance and shape of the target object, MDNet struggles to maintain tracking, whereas DBF excels. The classifier updating strategy of MDNet every 20 frames is inadequate for rapidly changes, often mistaking the altered target object for the background. Conversely, DBF, despite updating the classifier at the same interval, incorporates a system model for additional estimation, effectively mitigating these challenges. The experimental results highlight that incorporating a system model into DBF substantially improves TIR object tracking effectiveness in complex scenarios. Fig.\ref{fig:failure} shows two tracking failure cases of DBF. The fast movement of the camera equipment in the \emph{cow} sequence aggravates the motion blur. At the same time, because the distractors in the background are entirely consistent in appearance and shape with the target object, the system model provides the wrong candidates to the observation model. In the \emph{head} sequence, the target object is occluded by distractors with the same appearance and motion information, causing misjudgment by the observation model.

\subsection{Discussions}\label{sec:4-5}

\subsubsection{Comparison with Transformer-based trackers}\label{sec:4-5-1}

\begin{table*}[t]\footnotesize
\centering
\caption{Comparison of the proposed DBF with three representative Transformer-based trackers.}
\label{tab:tran}
\begin{tabular}{lccccccc}
\toprule
\multirow{2}{*}{Trackers} &  & \multicolumn{3}{c}{LSOTB-TIR}        &  & \multicolumn{2}{c}{PTB-TIR} \\ \cline{3-5} \cline{7-8}
                         &  & Success & Precision & Norm. Precision &  & Success     & Precision     \\
\midrule
TransT~\cite{transt}         &  & 0.673        & 0.798          & 0.721               &  & 0.613            & 0.764              \\
MixFormer~\cite{mixformer}   &  & 0.660        & 0.778          & 0.703               &  & 0.612            & 0.753              \\
CSWinTT~\cite{ccwintt}       &  & 0.644        & 0.761          & 0.685               &  & 0.572            & 0.753              \\
\hline
DBF                          &  & 0.625        & 0.770          & 0.703               &  & 0.626            & 0.839              \\
\bottomrule
\end{tabular}
\end{table*}

Transformer-based trackers have shown excellent performance compared to CNN-based trackers and are no surprise dominating the visual object tracking community. We also compare our DBF with three recent representative Transformer-based trackers\footnote{To ensure fairness, all Transformer-based trackers were retrained on the LSOTB-TIR training set.}: TransT~\cite{transt}, MixFormer~\cite{mixformer}, and CCWinTT~\cite{ccwintt}, on the LSOTB-TIR and PTB-TIR benchmark datasets.

As shown in Table \ref{tab:tran}, DBF exhibits superior performance on the PTB-TIR dataset, achieving a success score of 0.626 and a precision score of 0.839, significantly outperforming indicating its closest competitor, TransT. Such a margin is a notable superiority in the accuracy of DBF locating the target object. We believe that DBF comprehensively surpasses Transformer-based trackers on the PTB-TIR dataset because the target objects in this dataset are all pedestrian, causing the difference between consecutive video frames to be small. The motion information provided by the system model makes up for the lack of attention mechanism and spatiotemporal data compared to the Transformer architecture. In contrast, on LSOTB-TIR, a dataset with more object classes and challenging scenarios, Transformer-based trackers outperformed DBF. However, DBF maintains competitive precision and normalized precision scores, even outperforming CSWinTT. It is worth noting that while DBF does not lead on the LSOTB-TIR dataset, its consistent performance across different datasets for both success and precision metrics suggests robustness. This could imply that DBF has a more generalized model, which is less susceptible to dataset-specific biases or overfitting. Inspired by this, we will explore Transformer-based observation models to further improve the accuracy while maintaining or enhancing the robustness of DBF.

\subsubsection{Motion distributions}\label{sec:4-5-2}

\begin{figure*}[htb]
\begin{center}
    \subfigure{\includegraphics[width=.4\linewidth]{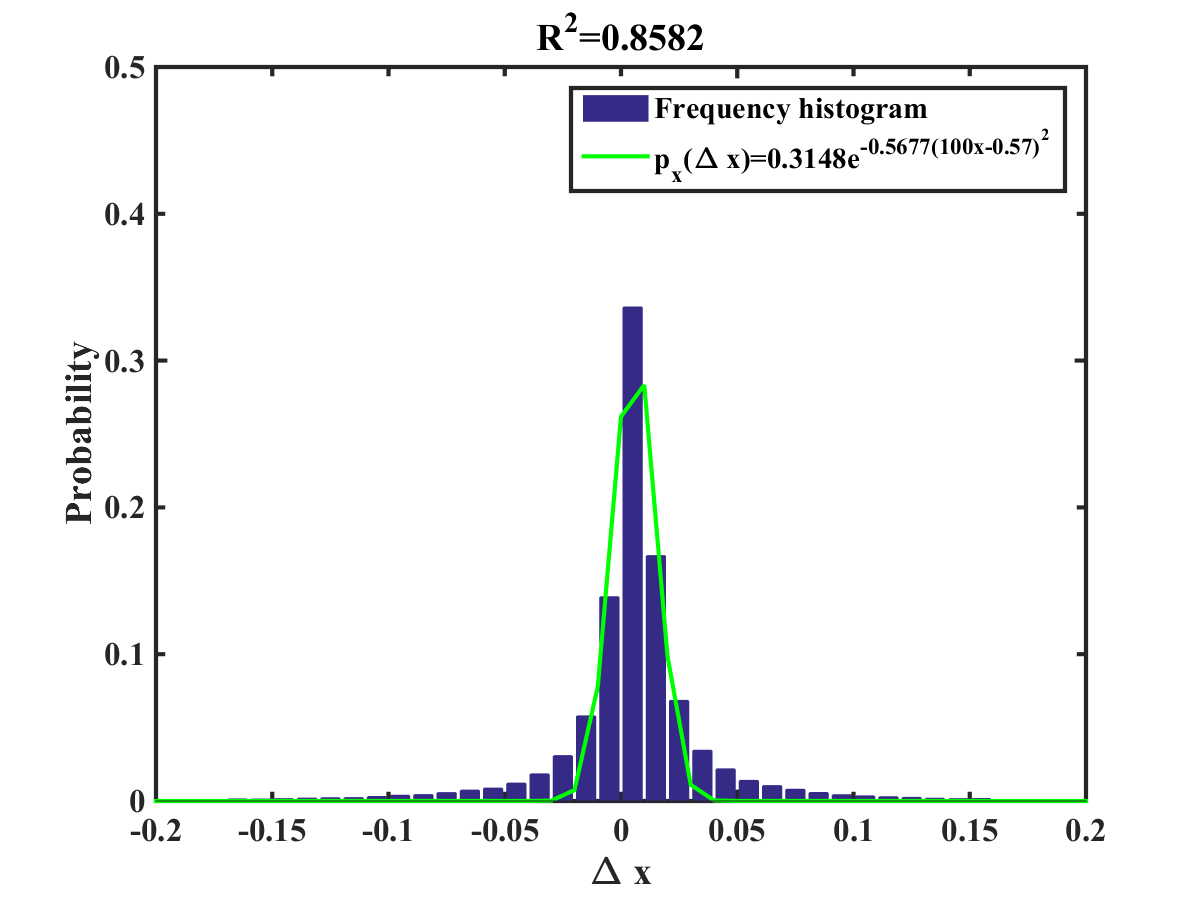}}
    \hspace{0.05em}
    \subfigure{\includegraphics[width=.4\linewidth]{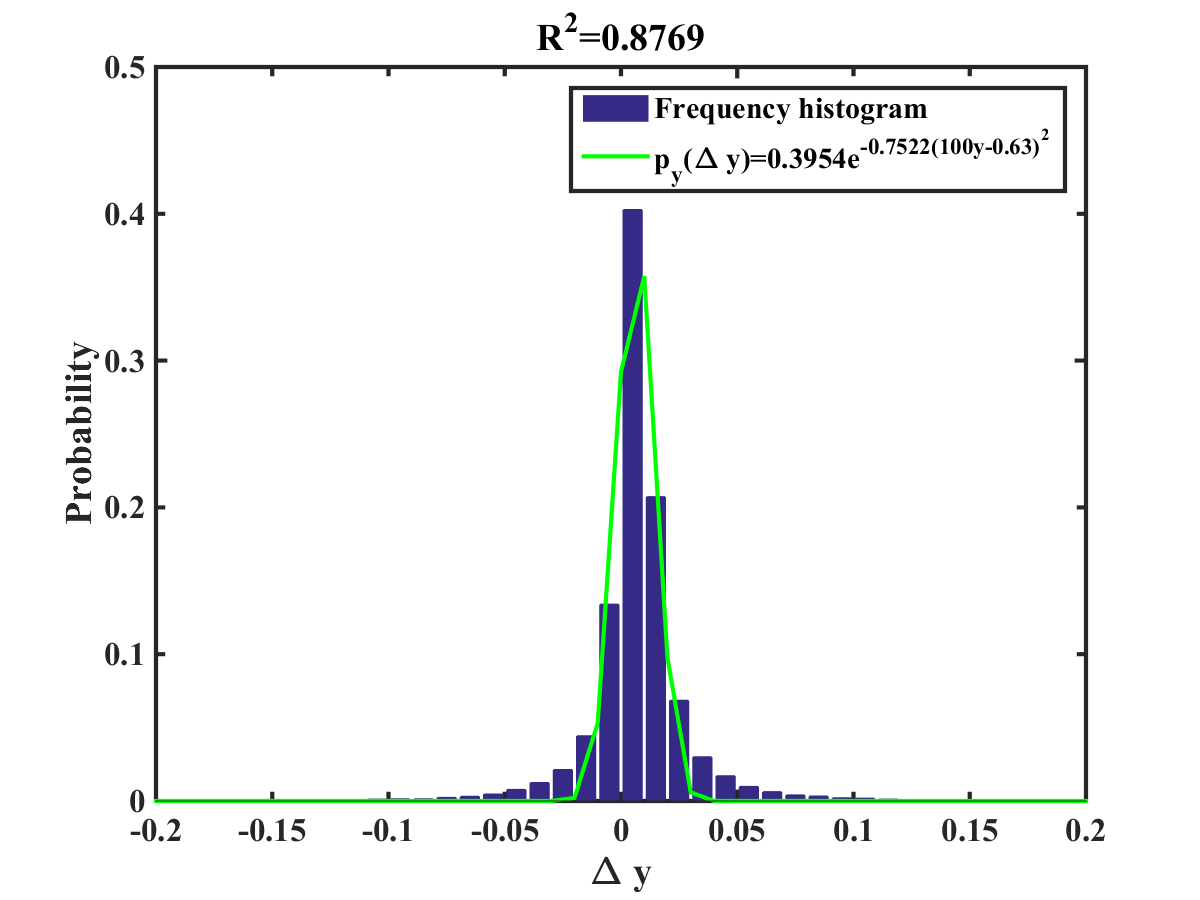}}
    \hspace{0.05em}
    \vfill
    \subfigure{\includegraphics[width=.4\linewidth]{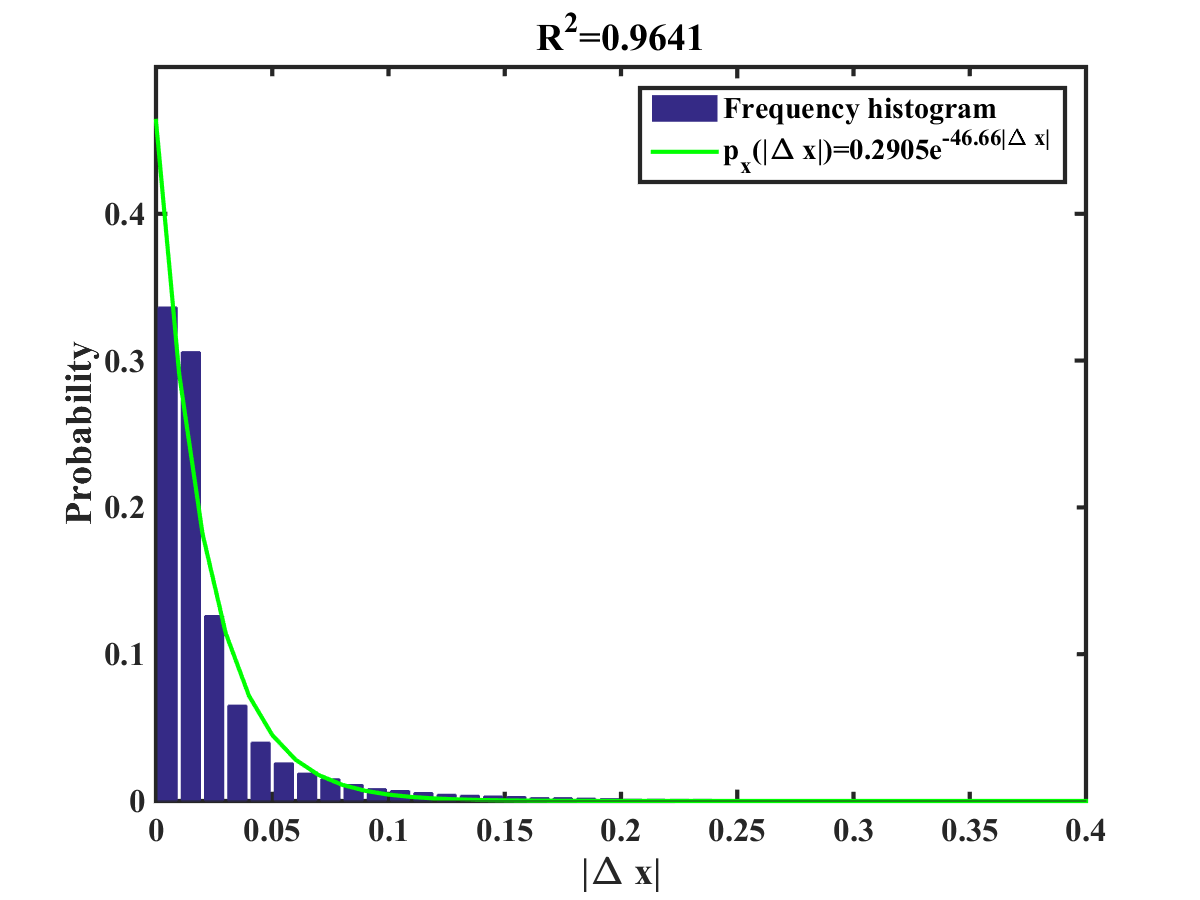}}
    \hspace{0.05em}
    \subfigure{\includegraphics[width=.4\linewidth]{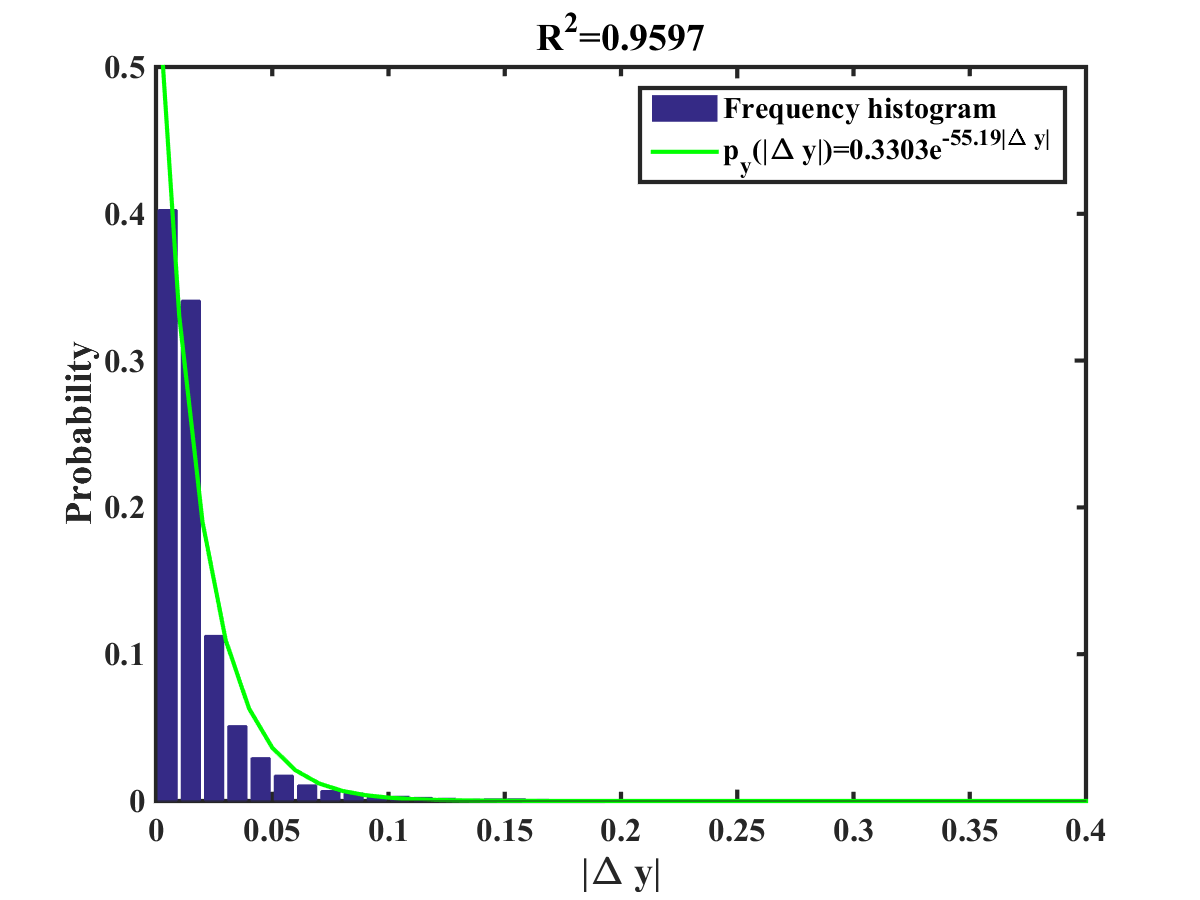}}
\end{center}
\vspace{-0.8em}
\caption{Evaluation of four scenario attribute subsets of the LSOTB-TIR~\cite{lsotb} benchmark dataset.}
\label{fig:dis}
\end{figure*}

In Section~\ref{sec:3-3}, based on the assumption of Brownian motion, we use a Gaussian distribution with zero means to describe the object motion information. In order to obtain a better system model, we further studied the motion distribution of target objects in the LSOTB-TIR and PTB-TIR benchmark datasets. These two datasets contain a total of 111,982 frames. We need to convert the ground-truth annotations of these target objects into the system state, that is, the position difference $(\Delta x, \Delta y)$ of the target object in two consecutive frames can be calculated as:
\begin{equation}
    \left\{
        \begin{aligned}
            &\Delta x=\frac{x_t-x_{t-1}+\frac{1}{2}(w_t-w_{t-1})}{\frac{1}{2}w_{t-1}+\frac{1}{2}w_{t}},\quad t>1  \\
            &\Delta y=\frac{y_t-y_{t-1}+\frac{1}{2}(h_t-h_{t-1})}{\frac{1}{2}h_{t-1}+\frac{1}{2}h_{t}},\quad t>1
        \end{aligned}
    \right.
\end{equation}
where $x_t$, $y_t$, $w_t$, and $h_t$ represent the top-left abscissa, the top-left ordinate, the width, and the height of the ground-truth annotation, respectively. Since the scale of the target object changes over the TIR video sequence, we use the relative motion of the centroid of the ground-truth annotation to represent the position difference $(\Delta x, \Delta y)$ of the target object. Fig.\ref{fig:dis} shows the histogram after converting 111,982 ground-truth annotations into position differences of the target object. We fit the frequency histograms of $\Delta x$ and $\Delta y$ using a Gaussian distribution --- the topper two graphs of Fig.\ref{fig:dis} show fitting results.

It is clear that the histogram appears to approximate a Gaussian distribution, indicating the Gaussian distribution's ability to effectively characterize motion differences. The density center of the histogram is near zero, and the mean values of the best Gaussian distribution fits for $\Delta x$ and $\Delta y$ have mean values of 0.0057 and 0.0063, respectively. This observation ratifies the zero-mean hypothesis of the Gaussian distribution. However, the peak of the histogram is much sharper than that of the Gaussian distribution, leading to a suboptimal fit. This discrepancy prompts us to explore an alternative distribution with a sharper peak, which might offer a better fit for the histogram. The Laplace distribution, with its sharper peak at the same mean, is more appropriate for fitting the absolute values of the position differences $|\Delta x|$ and $|\Delta y|$, given its domain of $[0,+\infty)$. The lower pair of graphs in Fig.\ref{fig:dis} illustrates the fitting of the Laplace distribution for $|\Delta x|$ and $|\Delta y|$. It is clear from these graphs that the Laplace distribution, for both $|\Delta x|$ and $|\Delta y|$, achieves a higher correlation coefficient R2 than that of the Gaussian distribution. In statistical analysis, R2 is employed to assess the accuracy of a fitted curve in representing actual data. Therefore, based on these results, we adopt the Laplace distribution over the Gaussian distribution to model the motion distribution $p(s_t|s_{t-1})$ of the target object as:
\begin{equation}
    \left\{
        \begin{aligned}
            &p(s_t|s_{t-1})=p_x(\Delta x)p_y(\Delta y)  \\
            &p_x(\Delta x)=\frac{\lambda_x}{2}e^{-\lambda_x|\Delta x|}  \\
            &p_y(\Delta y)=\frac{\lambda_y}{2}e^{-\lambda_y|\Delta y|}
        \end{aligned}
    \right.
\end{equation}
where $\lambda_x>0$ and $\lambda_y>0$ are the conversion coefficients.

\begin{table*}[t]\footnotesize
\centering
\caption{Comparison of using Gaussian distribution and Laplace distribution as system models of DBF. The larger the value, the better the performance.}
\label{tab:dis}
\begin{tabular}{lccccccc}
\toprule
\multirow{2}{*}{System Model} &  & \multicolumn{3}{c}{LSOTB-TIR}        &  & \multicolumn{2}{c}{PTB-TIR} \\ \cline{3-5} \cline{7-8}
                         &  & Success & Precision & Norm. Precision &  & Success     & Precision     \\
\midrule
Gaussian    &  & 0.625   & 0.770     & 0.703           &  & 0.626       & 0.839              \\
Laplace     &  & 0.633   & 0.796     & 0.737           &  & 0.641       & 0.862              \\
\bottomrule
\end{tabular}
\end{table*}

The effectiveness of system models based on Gaussian and Laplace distributions was compared using the LSOTB-TIR and PTB-TIR benchmark datasets, with the results detailed in Table \ref{tab:dis}. The results clearly show that the system model based on the Laplace distribution outperforms the one using the Gaussian distribution, demonstrating the effectiveness of our improvement in the system model.

\section{Conclusion}\label{sec:5}

This study proposes a novel method, DBF, that revisits and integrates Bayesian filtering with deep learning techniques for TIR object tracking. It enables the integration of motion data independent of infrared information, thereby enhancing the tracker's ability to predict and distinguish rapid changes in the appearance of the target objects. This is particularly crucial in scenarios where infrared information alone may falter. Moreover, Bayesian filtering introduces a probabilistic model updating strategy. This adaptive strategy significantly reduces the risk of tracking errors and drifts. Experimental evaluations on multiple benchmark datasets confirm that DBF outperforms most existing TIR tracking methods in complex scenarios, thus not only vindicating but also revitalizing the role of Bayesian filtering in this domain. The successful implementation of DBF paves the way for more robust and effective TIR tracking technologies capable of overcoming the nuanced challenges posed by intricate tracking environments. However, there are currently the following shortcomings: (a) the classifier used by the observation model is relatively simple, which is insufficient for TC and DIS cases with small differences in object classes, (b) although motion information can serve as spatiotemporal correlation between frames, it is still not sufficient in case of FM, and (c) competitive tracking performance has not been achieved in case of LR. In future work, we would like to employ more powerful network architectures from off-the-shelf Transformer-based and segmentation-based visual object tracking methods as the classifier of the observation model and adopt super-resolution methods to provide more robust TIR tracking performance.

\balance

\bibliographystyle{IEEEtran}
\bibliography{manuscript}

\end{document}